\documentclass[10pt,twocolumn,letterpaper]{article}

\usepackage{style/3dv}
\usepackage{times}
\usepackage{epsfig}
\usepackage{graphicx}
\usepackage{amsmath}
\usepackage{amssymb}
\usepackage[font=small]{caption}
\usepackage{xr-hyper}
\usepackage{subcaption}
\usepackage{gensymb}
\usepackage[percent]{overpic}
\usepackage{enumitem}
\usepackage{soul}

\usepackage[pagebackref=true,breaklinks=true,letterpaper=true,colorlinks,bookmarks=false]{hyperref}

\threedvfinalcopy 

\def\threedvPaperID{75} 

\begin{document}

\title{Reducing Drift in Structure From Motion Using Extended Features}

\author{Aleksander Holynski$^1{}^*$,   David Geraghty$^2$,    Jan-Michael Frahm$^2$,   Chris Sweeney$^3$,   Richard Szeliski$^2$ \\
$^1$University of Washington \qquad \qquad $^2$Facebook \qquad \qquad $^3$Facebook Reality Labs\\
}

\maketitle

\newcommand{\xvec}{\mathbf{x}} 
\newcommand{\xhat}{\mathbf{\hat{x}}} 
\newcommand{\pvec}{\mathbf{p}} 
\newcommand{\oo}{\bm{\omega}} 
\newcommand{\Rmat}{\mathbf{R}} 
\newcommand{\Imat}{\mathbf{I}} 
\newcommand{\crossMat}[1]{[\ #1\ ]_\times} 
\newcommand{\cvec}{\mathbf{c}} 
\newcommand{\tvec}{\mathbf{t}} 
\newcommand{\that}{\mathbf{\hat{t}}} 
\newcommand{\nvec}{\mathbf{n}} 
\newcommand{\nhat}{\mathbf{\hat{n}}} 
\newcommand{\lvp}{\lambda_{\mbox{vp}}} 
\newcommand{\Wvp}{W_{\mbox{vp}}} 
\newcommand{\dhat}{\hat{d}} 
\newcommand{\degr}[0]{${}^\circ$} 

\setlength{\abovedisplayskip}{2pt}
\setlength{\belowdisplayskip}{2pt}

\begin{abstract}
\vspace{-0.5em}
Low-frequency long-range errors (drift) are an endemic problem in 3D structure from motion, and can often hamper reasonable reconstructions of the scene. In this paper, we present a method to dramatically reduce scale and positional drift by using extended structural features such as planes and vanishing points. Unlike traditional feature matches, our extended features are able to span non-overlapping input images, and hence provide long-range constraints on the scale and shape of the reconstruction. We add these features as additional constraints to a state-of-the-art global structure from motion algorithm and demonstrate that the added constraints enable the reconstruction of particularly drift-prone sequences such as long, low field-of-view videos without inertial measurements. Additionally, we provide an analysis of the drift-reducing capabilities of these constraints by evaluating on a synthetic dataset. Our structural features are able to significantly reduce drift for scenes that contain long-spanning man-made structures, such as aligned rows of windows or planar building facades.
\end{abstract}
\makeatletter{\renewcommand*{\@makefnmark}{}
\footnotetext{$^*$This work was done while Aleksander was as an intern at Facebook.}\makeatother}
\makeatletter{\renewcommand*{\@makefnmark}{}
\footnotetext{Author email address: {\tt holynski@cs.washington.edu}.}\makeatother}
\section{Introduction} 
Structure from Motion (SfM)~\cite{Schonberger2016} and
Simultaneous Localization and Mapping (SLAM)~\cite{Fuentes2015, Engel2014, Engel2018} algorithms
serve as the foundation for a wide variety of applications in computer vision, including 3D reconstruction \cite{Agarwal2009}, 3D exploration of photo collections \cite{Snavely2006}, and phone-based augmented reality \cite{Schops2014}.
Given a set of images as input, SfM and SLAM systems reconstruct the per-image camera locations and orientations, as well as a sparse set of 3D points.

Thanks to remarkable speed and accuracy improvements over the last decade \cite{Engel2018}, these camera-based tracking methods can be used to quickly reconstruct the layout or 3D model of a scene, simply by capturing a short video clip from multiple viewpoints. Creating an accurate model, however, requires highly accurate poses; but despite recent improvements, today's best algorithms still suffer from long-range \emph{drift}, which results from the accumulation of small estimation errors caused by noise in feature point location estimates and other unmodelled sources of error. Significant drift can result in bent or deformed reconstructions, such as the one shown in Figure~\ref{fig:teaser}a.
\newlength\foa
\setlength\foa{4.15cm}
\begin{figure}[!t]

\centering%
\parbox[t]{\foa}{\vspace{0cm}\centering%
  \includegraphics[angle=270,origin=c,width=\foa]{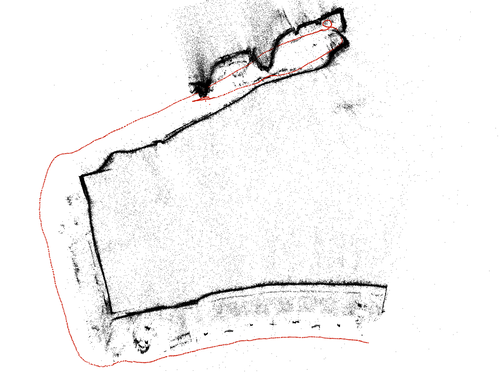}\\%
	\small (a) Standard structure from motion}%
\hfill%
\parbox[t]{\foa}{\vspace{0cm}\centering%
  \includegraphics[angle=270,origin=c,width=\foa]{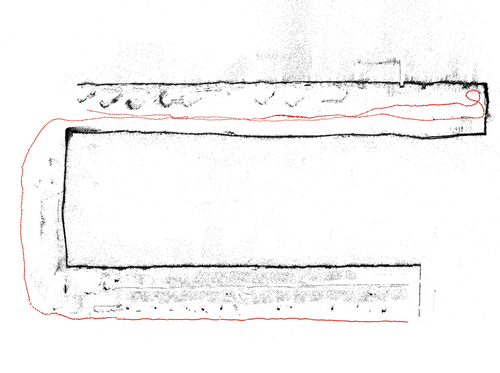}\\%
	\small (b) Our method, with added extended feature constraints}%
\caption{Standard SfM suffers from significant rotation and scale drift. In scenes with long-spanning structures, like an aligned row of windows wrapping around a building, or a long planar surface, our method can correct for this drift.}
\vspace{-0.5cm}
\label{fig:teaser}
\end{figure}

While all reconstructions contain some amount of (sometimes unnoticeable) drift, certain camera configurations are significantly more susceptible. For instance, narrow field-of-view cameras, like those in mobile phones, are much more prone to drift, since features persist for fewer frames when the camera is in motion. Since these feature tracks are the basis of motion estimation, if they are not observed from significantly different viewpoints, there are no direct constraints on the relative poses of those frames, and thus nothing to counteract the accumulation of drift error.

In fact, when trying to reconstruct objects which span a space larger than the field-of-view of the camera (such as a building or office-space), a drift-free reconstruction may be nearly impossible to capture. Those familiar with the limitations of these systems might attempt to choose camera paths that result in longer feature tracks, by keeping a greater distance from the subject, or fixating on single points; but often such paths are not feasible. For instance, to capture a building that is surrounded by trees, we're forced to capture the building from a very small distance. In such cases, our best option is to face the object and trace its perimeter, but doing so can result in significant drift (Fig.~\ref{fig:teaser}a).

In this paper, we demonstrate how tracking \emph{extended virtual features} such as co-planar sets of points, which span non-overlapping image frames or sometimes even complete video sequences, can dramatically reduce accumulated drift, enabling us to reconstruct sequences such as those described above. We focus on extended features which are \emph{structural}, namely vanishing points and oriented planes, which are common in man-made environments, such as cities and buildings. Note that our system does \emph{not} make Manhattan World assumptions, i.e., that all dominant structural planes are orthogonal. It can handle Atlanta World scenes with structures at arbitrary orientations \cite{schindler2004atlanta}.

To demonstrate the effectiveness of such features, we incorporate them into Theia \cite{Sweeney2015}, a popular open-source SfM system, and evaluate on both synthetic and real-world captures of scenes with strong planar structures using low field-of-view cameras. We show that the addition of extended features drastically reduces drift in both cases, producing results that surpass in quality both Theia and COLMAP, two of the leading open-source SfM packages.

We show the utility of extended features applied to a global SfM framework. Global SfM systems, in contrast to \emph{incremental} systems, solve for all camera poses at once, and are typically much faster, albeit less robust to errors in pairwise pose and correspondence. We show that for our drift-prone sequences, adding extended features to a global SfM system can produce higher quality reconstructions than an incremental SfM system at a fraction of the processing time. This is possible because our modifications preserve the efficiency of previous global methods, only adding a small number of extra parameters and a linear number (in the number of cameras) of additional constraints.

We begin the paper in Sec.~\ref{sec:previous} with a review of the previous literature,
followed in Sec.~\ref{sec:triplets} with a description of our global SfM baseline.
Sec.~\ref{sec:structural} then describes the structural components of our algorithm,
including the use of vanishing points for orientation estimation, local
orthogonal plane fitting, and the integration of these planes into the global position solver. 
Sec.~\ref{sec:evaluation} presents our experimental results on both synthetic data and real-world mobile phone captures. We then wrap up with a discussion of our results.

\section{Previous work}
\vspace{-0.4em}
\label{sec:previous}
SfM and SLAM methods generally consist of the same set of components:
identifying correspondences, pairwise pose estimation, and global integration.%
\makeatletter
\renewcommand{\paragraph}{%
  \@startsection{paragraph}{4}%
  {\z@}{0.8ex \@plus 1ex \@minus .2ex}{-1em}%
  {\normalfont\normalsize\bfseries}%
}
\makeatother
\paragraph{Correspondence.}
Both SfM and SLAM methods begin by establishing correspondences, which capture the motion observed between pairs of frames and serve as the foundation for pose estimation. 
Traditionally, the features used for correspondence are 2D point descriptors \cite{Harris1988, Lowe1999, Tuytelaars2008}. These descriptors analyze local image patches to determine salient keypoints and are invariant to geometric and radiometric transformations. Other types of features include line segments \cite{Baillard1999, Rother2003, Micusik2017}, vanishing points \cite{Sinha2010}, 3D planes \cite{Debevec1996, Lang1996, Rother2003, Bartoli2000, Bartoli2003}, and inertial measurements \cite{LiMouriks-ijrr2013}. 
\paragraph{Pairwise pose estimation.}
Once the correspondences have been established and filtered,
two-view or multi-view geometry estimation techniques \cite{Nister2004, Pollefeys2002, Haralick1989, Faugeras1992, Hartley2003, Hartley1997, Torr1997} are used to recover relative camera pose between pairs of images, which, along with point correspondences, can be used to triangulate 3D points.  These pairwise relative camera poses and 3D points are effectively \emph{local} reconstructions consisting of two or three frames. 
\paragraph{Global integration.}
\label{sec:global_integration}
In order to produce a global reconstruction,
it is necessary to combine the local pose estimates.
Techniques for robustly integrating these local reconstructions are usually divided into two categories: incremental and global. 

Incremental methods~\cite{Snavely2010, Wu2011, Schonberger2016, Agarwal2009, Frahm2010, Sweeney2015} have long dominated the state of the art, and can be found in the majority of open-source implementations, such as Bundler \cite{Snavely2010}, VisualSfM \cite{Wu2011},
and COLMAP \cite{Schonberger2016}.
 This form of global integration typically starts with a carefully selected two or three-frame reconstruction~\cite{Snavely2006, Snavely2010, Schonberger2016}, and incrementally grows it by selecting and anchoring new frames to the already-reconstructed cameras and points. To account for the inconsistencies from the newly registered poses, each frame addition is typically followed by a number of filtering operations to verify and refine the pose
 \cite{Snavely2006,Schonberger2016}.
 While these filtering operations cause the method to be very robust to outliers, they can also be very computationally expensive. 
To improve efficiency, incremental systems employ a number of optimizations to reduce incurred processing time \cite{Bhowmick2014, Steedly2003, Agarwal2009, Wu2011_CVPR, Ni2007}. SLAM methods~\cite{Engel2014, Mur2015, Engel2018, Qin2018, Forster2016} also belong to this category, as they incrementally track the camera pose by accumulating differential camera motion.

Global methods \cite{Govindu2001, Rother2003, Sinha2010, Jiang2013, Moulon2013, Wilson2014, Micusik2017} estimate all camera poses simultaneously, making them efficient for large-scale problems. These methods are generally regarded as being less robust to outliers, as the lack of an incremental reconstruction precludes the ability to identify pairwise pose outliers by verifying against a global model. 
To improve robustness, global SfM methods utilize either groups of frames \cite{Moulon2013, Jiang2013, Govindu2006}, observed 3D landmarks \cite{Sinha2010, Cui2015}, or pairwise geometry analysis \cite{Steedly2003} to further verify pairwise pose estimates before integrating globally. 

\paragraph{Drift Mitigation.} 
Incremental and global techniques both suffer from drift, 
particularly in long sequences without loop closures,
caused by the accumulation of small relative pose errors. A number of avenues have been explored to mitigate these errors.

SfM systems will often perform global bundle adjustment \cite{Triggs1999} over all reconstructed frames and points. This process can reduce, {\em{but not eliminate}}, drift error, since the bundle-adjusted reconstruction can only be as good as the correspondences used for optimization. If there are correspondence errors,
the resulting reconstruction will inevitably contain some amount of drift. 

Real-time SLAM methods~\cite{Engel2014, Mur2015, Engel2018, Qin2018, Forster2016} which cannot afford costly global bundle adjustment will often include inertial measurements \cite{LiMouriks-ijrr2013} as a secondary source of motion information. While these inertial sensors have become commonplace in modern mobile phones, most captured or distributed videos do not retain inertial measurements, and thus this information is often unavailable.

For closed-loop sequences, a common tactic to mitigate drift is to perform \emph{loop closure} or explicit matching and pairwise pose estimation between temporally distant frames that observe similar parts of the scene. While the constraints derived from these matches have the potential of reducing the drift in the reconstruction, in many cases, the drift error is simply redistributed to different parts of the reconstruction. This is because loop closure only adds constraints to a handful of images at the closure point, but does not apply direct constraints to the poses of images elsewhere. 

Higher-dimensional geometric features, like vanishing points, have been shown to significantly reduce rotational drift in both SLAM \cite{Camposeco2015} and SfM \cite{Sinha2010}, since they are able to extract direct constraints between pairs of frames that do not observe the same part of the scene. For active depth sensors, dense normal statistics \cite{Straub2014} provide a similar signal. These methods all provide strong constraints on rotational drift, but do not address \emph{translational} drift.

Other geometric features, such as 3D lines, have shown potential for reducing translational drift. Micusik \textit{et al}.~\cite{Micusik2017} incorporate line segment endpoints into an incremental SfM system, allowing accurate reconstruction in the absence of dense point features. Similarly, Zhou et al.~\cite{Zhou2015} propose a SLAM system that extracts and matches axis-aligned structural lines to apply constraints on the camera pose. Nurutdinova et al.~\cite{Nurutdinova2015} propose a generalization that includes arbitrary 3D curves in bundle adjustment. These methods rely on establishing correspondences between lines, either through endpoint matching or photometric comparison, techniques which, in real-world settings, are typically less reliable than point-based features. 

Planar constraints have also been shown to resolve translational and scale drift. Szeliski and Torr \cite{Szeliski1998} provide a theoretical introduction to the use of known planar structure as constraints in bundle adjustment. They show that for simple dataset of two cameras, the quality of reconstruction can be improved significantly by incorporating prior knowledge of planar scene structure into bundle adjustment. Extending upon this work, Rother \cite{Rother2003} proposed a factorization-based reconstruction system which jointly reconstructs camera pose, points, lines, and planes. Similar to \cite{Szeliski1998}, results are only shown on small scenes with few images, since factorization-based approaches are typically quite sensitive to outliers, and do not easily scale to large real-world datasets. Additionally, both methods require a known reference plane, unlike our method, which automatically discovers and associates structural elements. 

More recently, Liu et al.~\cite{Liu2016} demonstrated a SLAM framework that applies a piecewise-planar assumption in tracking, using homographies to efficiently track a camera under rapid motion. While the use of homographies enables fast and robust tracking, the method still suffers from significant drift over longer sequences, as planes are not associated across non-overlapping views. 

Li et al.~\cite{Li2018} show that additional constraints, such as coplanar sets of lines, found through vanishing point estimation and homography fitting, can further reduce positional drift. This method is reliant on long-spanning line segments in order to establish strong constraints, and requires mutually visible line endpoints for optimization. In contrast, our method does not require lines to be mutually visible in multiple views, and only relies on feature point correspondences, which are standard for SfM systems. Yang and Scherer~\cite{Yang2019} show that the boundaries between the semantic labels of surfaces such as walls, floor, and ceiling can also be used in constraining the camera position. This method relies on automatic semantic labeling, and thus does not easily extend to arbitrary planar structures. Cohen et al.~\cite{Cohen2015} show that many of these same assumptions, such as Manhattan-oriented structure, symmetry, and repeating elements, can facilitate the fusion of disconnected or sparsely overlapping reconstructions.

In this paper, we describe a technique that automatically extracts structural elements from a real-world video sequence, including vanishing points and planes, and automatically finds associations between non-overlapping observations of these elements in order to establish long-range constraints that reduce pose drift. Most similar to our work, Shariati et al.~\cite{Shariati2019} jointly optimize for wall positions relative to the camera, but require depth and inertial sensors, and also assume Manhattan structure. Our method takes as input only a monocular video sequence, and easily extends to scenes without a global Manhattan coordinate frame.

\begin{figure*}[t]

\centering%
\vspace{0cm}\centering%
  \includegraphics[width=\textwidth]{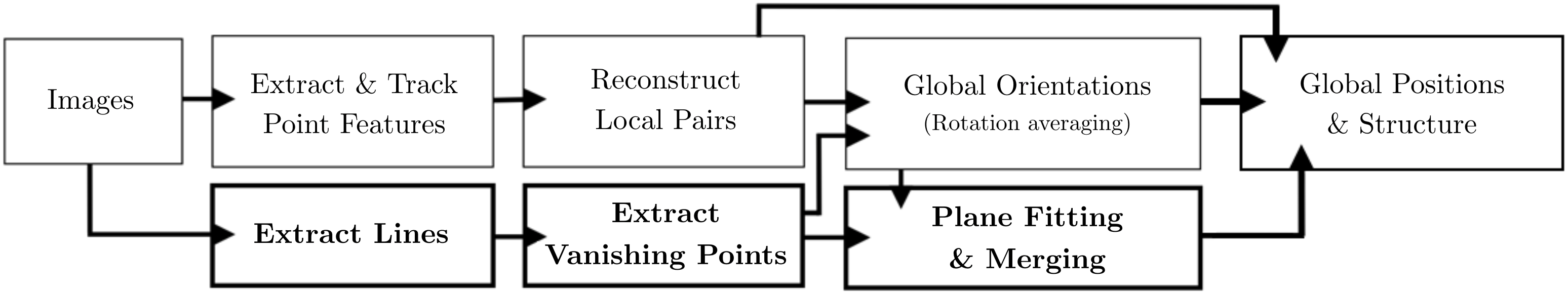}\\%

\caption{An overview of the global SfM system, showing the baseline components (non-bold) and our added structural constraints (bold).
}%
\vspace{-0.3cm}
\label{fig:pipeline}
\end{figure*}

\section{Baseline global SfM pipeline}
\label{sec:triplets}
This section describes the baseline Theia system~\cite{Sweeney2015}, shown as non-bolded boxes in Fig.~\ref{fig:pipeline}. Section~\ref{sec:structural} describes our novel structural constraints, which are shown in bold. 
Theia is a point-based SfM system, using tracked point features to create local reconstructions, which establish consistent relative camera motion between pairs of frames. These local reconstructions are then used in rotation averaging \cite{Chatterjee2013} to estimate the global orientations of all the cameras (Sec.~\ref{sec:pipeline_global_rotation}). Finally, the global camera centers are computed using the estimated rotations (Sec.~\ref{sec:pipeline_global_position}), and the 3D points are triangulated to produce the final reconstruction. 

Theia implements a number of global rotation and position solvers. For our baseline system, we chose the  L1-IRLS rotation solver~\cite{Chatterjee2013} and the LUD position solver~\cite{Ozyesil2015}. These are the recommended default solvers in Theia and also consistently produced the best results on our datasets. 
Apart from two modifications, described in Sec.~\ref{sec:degenerate}, the implementation of Theia is unchanged from the publicly available library.
Here, we provide a quick review of the formulation used in both the rotation and position solvers.
\subsection{Problem formulation and notation}
\label{subsec:formulation}

We start with a set of \emph{point tracks} relating 2D point features
$\xvec_{ip}$ with camera (image) indices $i$ and 3D point indices $p$.
Since we assume known camera intrinsics, these points have already been
centered w.r.t.\ the image center and their pixel coordinates divided
by each camera's focal length $f_i$ so that the
$\xhat_{ip} = (x_{ip},y_{ip},1)$
correspond to metric (Euclidean) ray directions in each camera's frame.

The projection equations relating a 3D world coordinate $\pvec_p$ to
its corresponding 2D projection $\xvec_{ip}$ in image $i$ is then
\begin{equation}
    \xhat_{ip} \sim \Rmat_i (\pvec_p - \cvec_i) ,
\end{equation}
where $\Rmat_i$ is the camera's 3D orientation,
$\cvec_i$ is its 3D position, and
$\sim$ indicates similarity up to scale.

\subsection{Global orientations (rotation averaging)}
\label{sec:pipeline_global_rotation}
To solve for the global orientations of all cameras, global SfM systems typically use rotation averaging, which solves for the global camera orientations that best agree with pairwise rotation estimates by minimizing the consistency error
\begin{equation}
E_{\mbox{rot}}(\{\Rmat\}) = \sum_{i,j} || \Rmat_j^{-1}\Rmat_{ij}\Rmat_i - \Imat ||_1
\label{eqn:rotation_avg}
\end{equation}
where $\Rmat_i$, $\Rmat_j$ are the unknown global orientations of frames $i$ and $j$, respectively, and $\Rmat_{ij}$ is the known pairwise rotation estimate between the two frames.

We use Theia's robust L1-IRLS solver, proposed in \cite{Chatterjee2013}, which first performs an L1 minimization and then refines the solution using iteratively reweighted least squares.

\subsection{Global position estimation}
\label{sec:pipeline_global_position}

Once the global orientations of all cameras have been estimated, all pairwise constraints are then integrated to estimate global camera centers. Theia uses the "least unsquared deviation" (LUD) global position estimator proposed by \cite{Ozyesil2015}, which formulates the optimization as:
\begin{equation}
E_{\mbox{pos}}(\{\cvec,s\}) = \sum_{i,j} || s_{ij} \tvec_{ij} - \cvec_j + \cvec_i ||_2
    \label{eqn:global-unweighted}
\end{equation}
where $\tvec_{ij}$ is a known pairwise translation estimate (after rotating all cameras into the global coordinate system), $\cvec_i$ and $\cvec_j$ are the unknown camera centers, and $s_{ij}$ is the unknown scaling coefficient for the pairwise reconstruction. These equations are optimized using a fast convex solver.

\subsection{Extensions to handle degenerate configurations}
\label{sec:degenerate}
In order for the baseline Theia to work on our sequences (narrow field-of-view videos), which often contain
``degenerate cases'' such as linear camera motions and single planes (building walls),
we added the following extensions, which are described in more detail in our supplementary material:
\begin{enumerate}[nolistsep,leftmargin=0.5cm]
    \item The above global position estimation suffers from degeneracy for colinear camera motion (as described in \cite{Jiang2013, Moulon2013}). In order to resolve this, we adapt a simplified version of \cite{Moulon2013}, applying additional constraints on the relative scales of pairwise reconstructions by comparing the triangulated depths of shared feature tracks.
    \item  In order to deal with planar scenes and pure rotations, for which the 5-point algorithm may produce degenerate configurations, we additionally estimate pairwise pose from a homography (using \cite{Malis2007}), keeping whichever approach produces a larger number of inliers.
\end{enumerate}

\section{Structural constraints}
\label{sec:structural}

In order to reduce the drift (global low-frequency errors and deformations)
in our reconstructions, we exploit large-scale \emph{structural constraints}
such as vanishing points and planes.
These can be thought of as \emph{extended features} since they will often span
many more frames than traditional point tracks, which come in and out
of view. 

\begin{figure}
\hfill
  \begin{overpic}[trim=0 0 0 80,clip,width=0.32\linewidth]{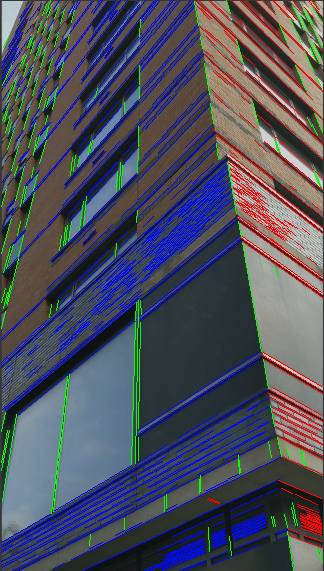}
  \end{overpic} \hfill
    \begin{overpic}[trim=0 0 0 80,clip,width=0.32\linewidth]{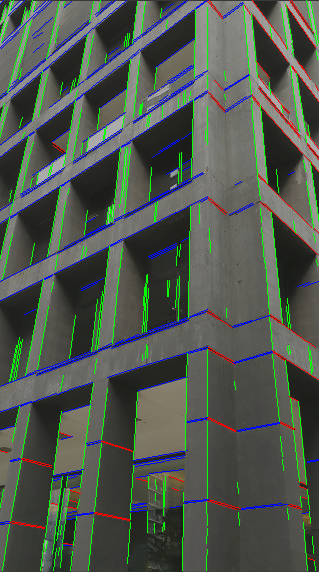}
  \end{overpic} \hfill
      \begin{overpic}[trim=0 0 0 80,clip,width=0.32\linewidth]{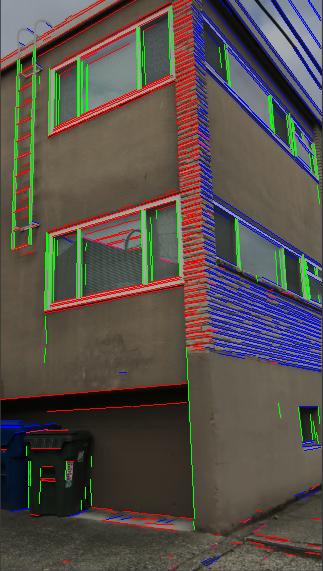}
  \end{overpic} \hfill
  \caption{Examples of our edge detection and vanishing point estimation stage. Different colored edges indicate their membership in different vanishing points.
  Note that only two orthogonal vanishing points need to be detected to establish the coordinate frame. \vspace{-0.4cm}}
    \label{fig:vps}
\end{figure}

\subsection{Lines and vanishing points}
\label{subsec:vanishing points}
In order to obtain a drift-free set of rotation estimates, we first compute for each frame (wherever possible) a vertical vanishing point and one or more
horizontal vanishing points, as described in the supplemental material
and illustrated in Fig.~\ref{fig:vps}.
Once the vanishing points have been found in frame $i$, we estimate a global rotation $\Rmat_i^{\mbox{vp}}$, which maps one of the Atlanta world horizontal directions to the dominant horizontal direction in the frame. Compared to the previous work by Sinha \emph{et al.}~\cite{Sinha2010}, which performs a global matching step between vanishing directions of all frames, our approach is tailored for
continuous video sequences where motion between frames is small, so the vanishing point associations can be chained through consecutive frames. 
This gives us a consistent \emph{soft} (drift-free, but empirically lower-accuracy) global orientation constraint for every frame that has associated vanishing point data.

To take advantage of these estimated points, we initialize the solution for the global camera orientations using the vanishing points,
 $    \Rmat_i \xleftarrow{} \Rmat_i^{\mbox{vp}}$.
In addition to the inter-frame constraints from pairwise pose estimates \eqref{eqn:rotation_avg}, we also add penalties on the difference between the current rotation estimates and the vanishing-point based rotations:
\begin{equation}
  E_{\mbox{vp}}(\{\Rmat\}) = \lvp \sum_i  \Wvp(i)
  \| \Rmat_i^{-1} \Rmat_i^{\mbox{vp}} - \Imat \|_1 .
\end{equation}
These constraints are weighted using a regularization parameter $\lvp$ and a per-frame weighting function
\begin{equation}
\Wvp(i) = \text{clamp}(1 - \frac{\Delta \theta}{\theta_{max}}, 0, 1)
\end{equation}
where $\Delta \theta$ is the incremental rotation between frames $i-1$ and $i$ and $\theta_{max}$ is the largest tolerated incremental rotation. We use $\theta_{max}=5$ degrees for all experiments. This weighting function makes the rotation estimates more robust to significant outliers in the vanishing point estimates, by lowering the weight of consecutive frames when they are very different. Further information on vanishing point estimation and integration is provided in the supplementary material.

\subsection{Extended planes}
\label{subsec:planes}
While adding vanishing points as soft constraints on camera orientations can dramatically
reduce orientation drift and increase reconstruction accuracy, positional drift still remains an issue. In order to address this, we exploit another major source of structural constraints, i.e., coplanar points
arising from man-made structures such as buildings.

In this section, we describe how we identify coplanar 3D points in local pairwise reconstructions and then link these together into extended global planes that can be
used as additional constraints in the pose estimation process.
As mentioned before, in our current system, we restrict our attention to planes whose normals correspond to one of the dominant vanishing point directions.

\paragraph{Local plane fitting.} We begin by discovering planes in each pairwise reconstruction. For each local reconstruction containing valid vanishing points, we use the pairwise pose estimate to perform two-view triangulation, resulting in a local 3D point cloud. We then perform a plane sweep along the three orthogonal vanishing directions associated with the base frame of the pairwise reconstruction. This results in a number of local planes $\pi$, parameterized as: \vspace{-1em}


\begin{equation}
    \pvec \cdot \nhat_\pi^{ij} = d_{ij}^\pi,
\end{equation}
where the $\pvec$ are the local 3D point inliers,
and $\nhat^{ij}_\pi$ is the local plane normal, and
$d_{ij}^\pi$ is the distance along the normal $\nhat$ from the origin
of pairwise estimate $i\xrightarrow{}j$. We provide more details on plane fitting in the supplemental material.

\paragraph{Plane merging and constraints.} Once local plane hypotheses have been generated for each pairwise reconstruction, we group these into global \emph{extended} planes, which we then use to 
provide additional constraints on the local scales and global camera
positions, as described below in Eq.~\ref{eqn:planar-unweighted}. In order to link these local hypotheses, we rely on point correspondences. Since local planes are established by finding co-planar sets of 3D points, each local plane contains a set of \emph{inlier} tracks which can be used to associate local planes with one another. We perform this association by greedily merging any two local planes which share a majority of tracks, and are associated with the same vanishing direction.

Once we have established estimates of which local plane hypotheses correspond to one another, we define constraints that encourage these local planes to coincide in the final 3D reconstruction. In order to integrate local plane hypotheses into global constraints on
camera positions, we add scalar variables $d_p$ to the linear system,
where $p$ is the index of the global extended plane (as opposed to the local plane index $\pi$).
These variables define each global plane's location
(distance to the world origin along the plane normal).
This corresponds to one added constraint for each observation of a global plane $p$ in a pairwise estimate $ij$:
\begin{figure}
\hfill
  \begin{overpic}[width=0.75\linewidth]{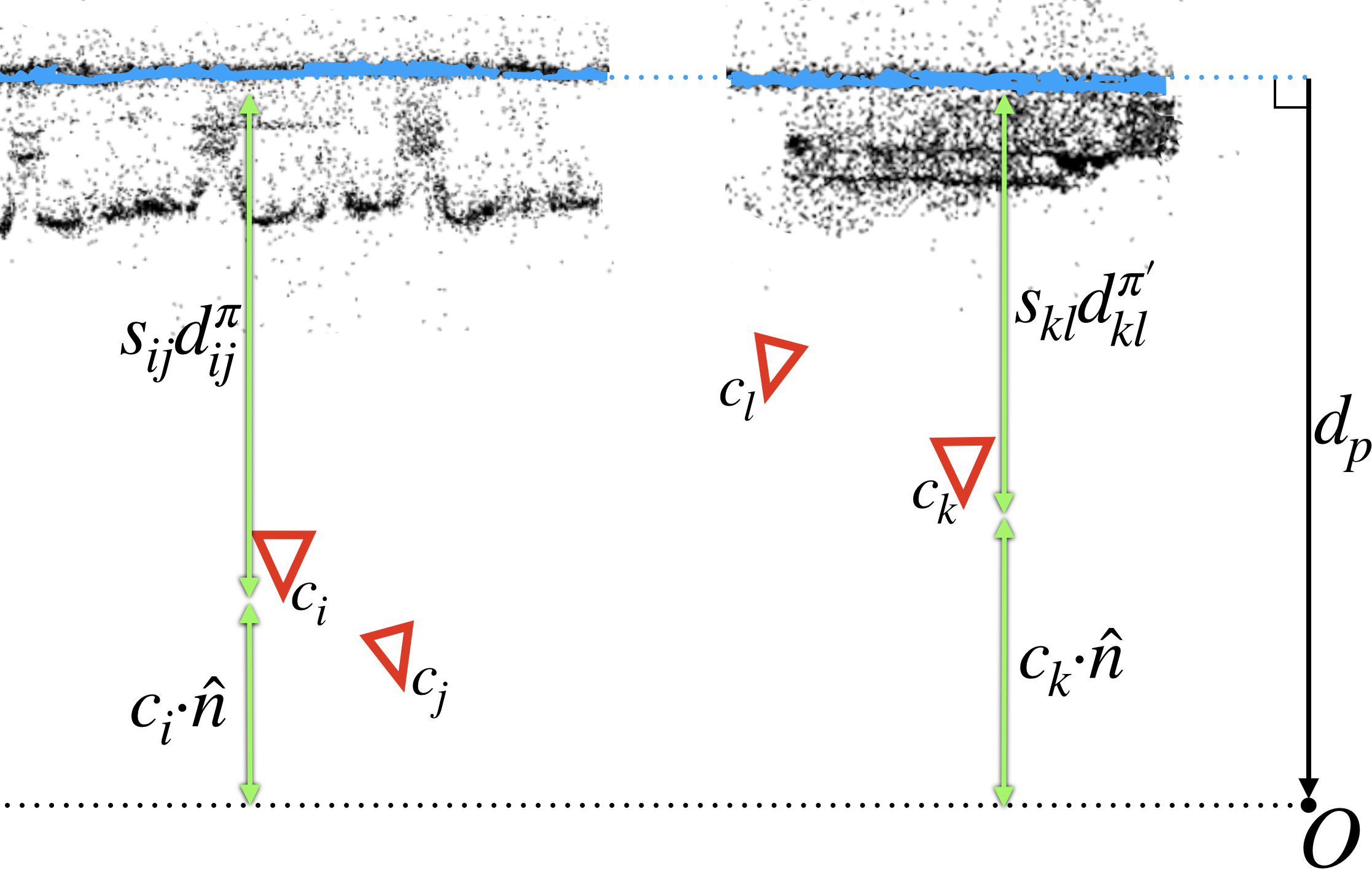}
  \end{overpic} \hfill
  \vspace{-0.3cm}
  \caption{An illustrated example of our plane constraints described in Eq.~\ref{eqn:planar-unweighted}. In the global coordinate frame, two pairwise local reconstructions $i\rightarrow j$ and $k\rightarrow l$ have their own local distances $d_{ij}^\pi$ and $d_{kl}^{\pi'}$ between the base cameras and their respective local planes $\pi$ and $\pi'$. When these planes correspond to the same global plane $p$ (blue), our plane constraints will align these two locally estimated planes in the global reconstruction by encouraging each local plane distance to be consistent with a single global distance from the origin $d_p$. This example shows the result of optimization, where our planar loss has been minimized such that $d_p = s_{kl}d^{\pi'}_{kl} + c_k\hat{n} = s_{ij}d^{\pi}_{ij} + c_i\hat{n}$, and $E_{\mbox{pln}} = 0$.}
  \vspace{-0.4cm}
    \label{fig:plane_formula}
\end{figure}

\begin{equation}
 E_{\mbox{pln}}(\{\cvec,s\}) = \sum_{i,j,p} W_{\mbox{p}}(ij, \pi) \|
    s_{ij} d_{ij}^\pi + \cvec_i \cdot \nvec_p - d_p \|_2
    \label{eqn:planar-unweighted}
\end{equation}
where the $s_{ij}$ is the unknown pairwise estimate scaling factor,
$\nvec_p$ is the known plane orientation (normal vector),
$\cvec_i$ is the unknown global camera position of the base camera in the pairwise transformation,
and $d_p$ is the unknown global distance of the plane from the world origin
along $\nvec_p$. Fig.~\ref{fig:plane_formula} provides a visualization of these terms.

The number of global planes $p$ (and hence extra scalar unknowns) is a small constant number per scene, and therefore the number of added constraints to the system is linear in the number of cameras. We weigh each of these constraints by the support of the global plane in the local pairwise reconstruction, i.e. by the number of inlier tracks. More formally, we define a weighting function 
\begin{equation}
    W_{\mbox{p}}(ij, \pi) = \min\left(\frac{S(ij, \pi)}{S_{\max}}, 1\right) * \lambda_{\mbox{p}}
\end{equation}
where $W_p(ij, \pi)$ defines the weight for plane $\pi$ in pairwise estimate $i\rightarrow{}j$,
$S(ij,\pi)$ returns the number of inliers in the local plane estimate, 
$S_{\max}$ defines the minimum number of inliers to receive full weight, and
$\lambda_{\mbox{p}}$ is the plane weighting coefficient (the maximum weight a plane can have). We use $S_{\max} = 10, \lambda_{\mbox{p}} = 50$ in all our experiments.
In Sec.~\ref{sec:evaluation}, we show how adding plane constraints dramatically improves reconstructions of both synthetic and real-world scenes. 

\section{Evaluation}
\label{sec:evaluation}
\vspace{-0.4em}

\begin{figure}
    \centering 
\begin{subfigure}{0.19\linewidth}
  \centering%
  \begin{overpic}[width=\linewidth]{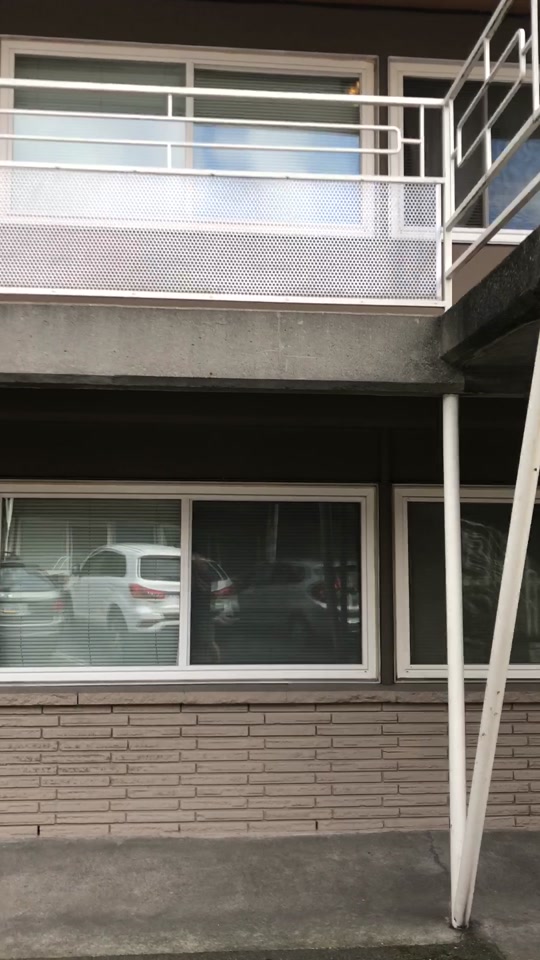}
  \end{overpic}
    \begin{overpic}[width=\linewidth]{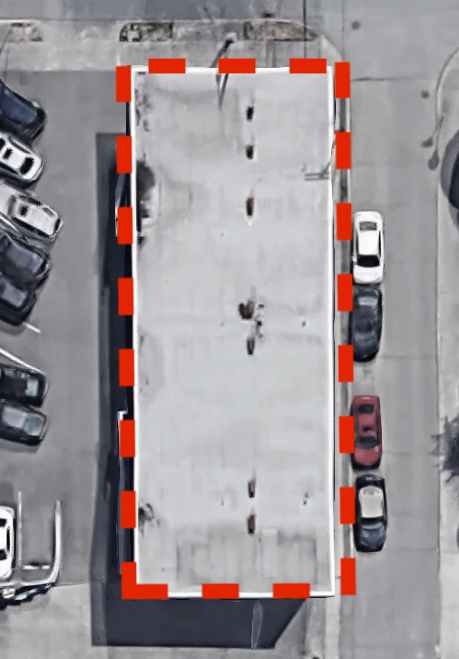}
  \end{overpic}
  \caption*{\small{seattle1}}
\end{subfigure}\hfil 
\begin{subfigure}{0.19\linewidth}
  \centering%
  \begin{overpic}[width=\linewidth]{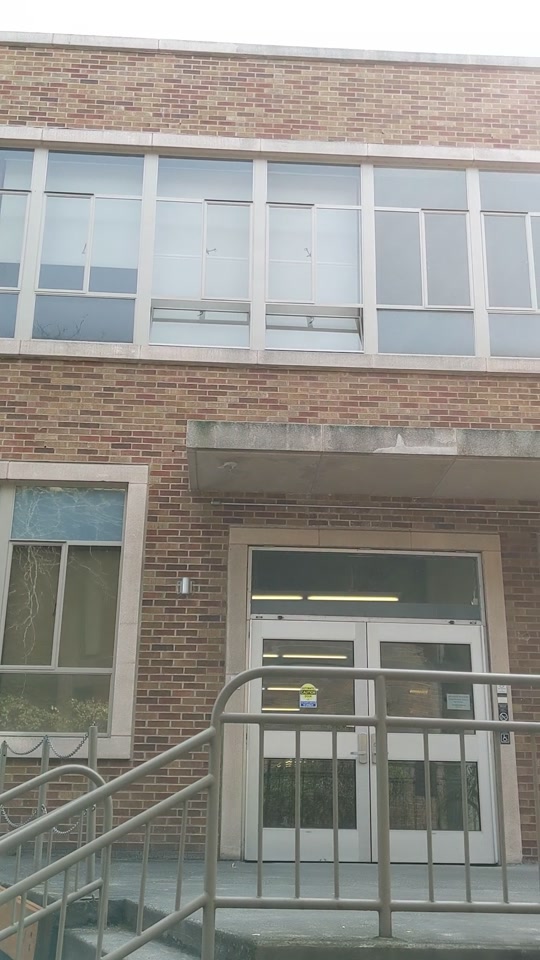}
  \end{overpic}
      \begin{overpic}[width=\linewidth]{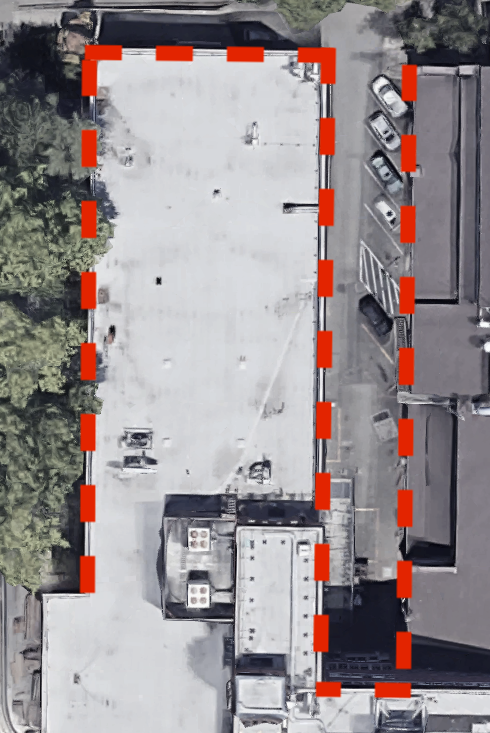}
  \end{overpic}
  \caption*{\small{more\_half}}
\end{subfigure}\hfil 
\begin{subfigure}{0.19\linewidth}
  \centering%
  \begin{overpic}[width=\linewidth]{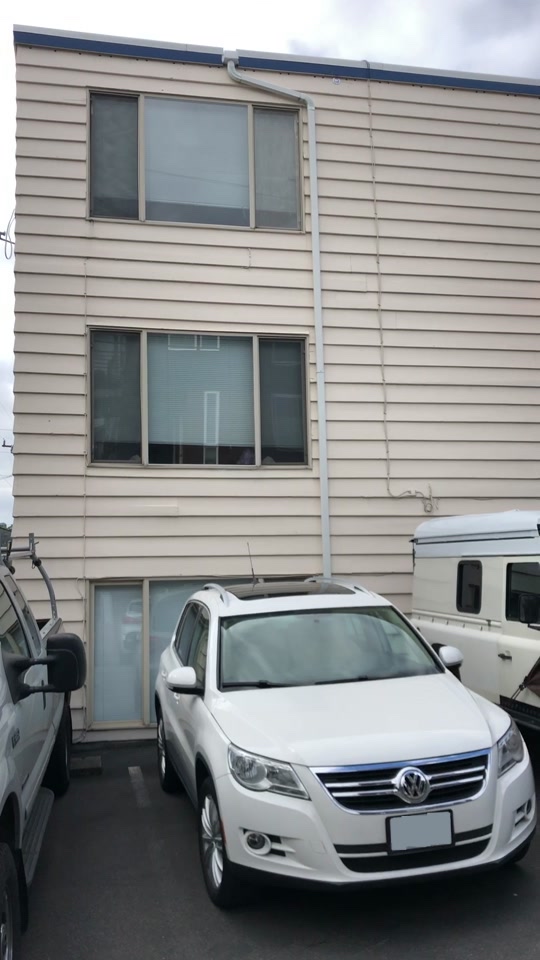}
  \end{overpic}
      \begin{overpic}[width=\linewidth]{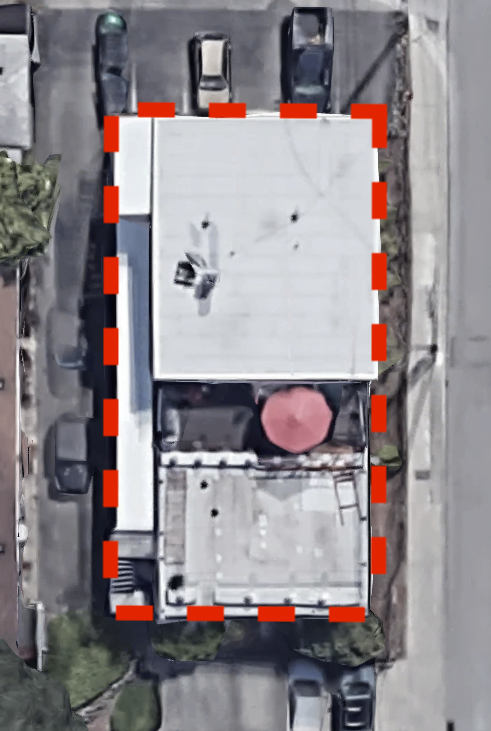}
  \end{overpic}
  \caption*{\small{seattle2}}
\end{subfigure}\hfil 
\begin{subfigure}{0.19\linewidth}
  \centering%
  \begin{overpic}[width=\linewidth]{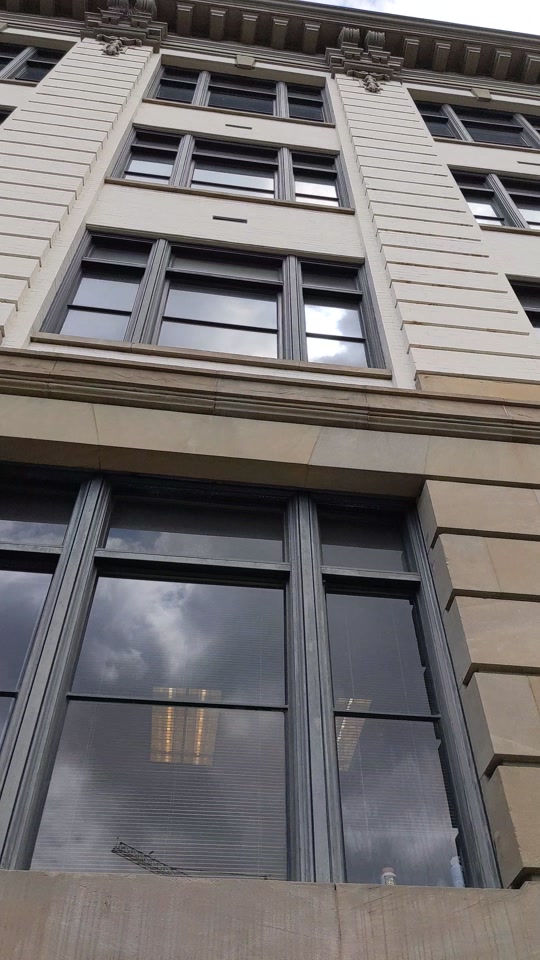}
  \end{overpic}
      \begin{overpic}[width=\linewidth]{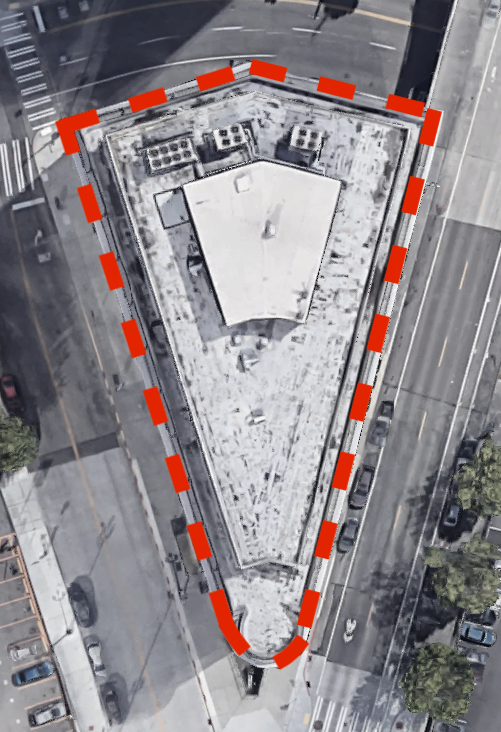}
  \end{overpic}
  \caption*{\small{atlanta1}}
\end{subfigure}\hfil 
\begin{subfigure}{0.19\linewidth}
  \centering%
  \begin{overpic}[width=\linewidth]{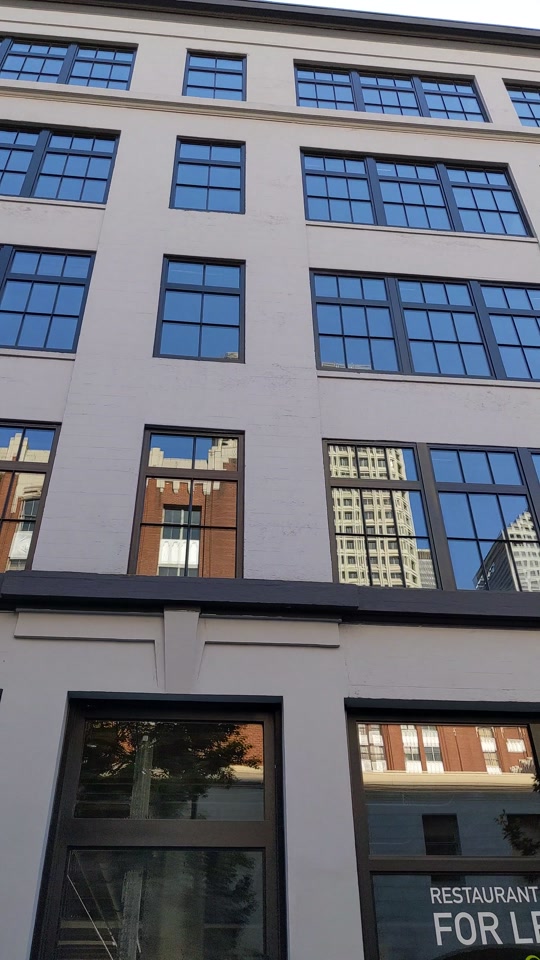}
  \end{overpic}
      \begin{overpic}[width=\linewidth]{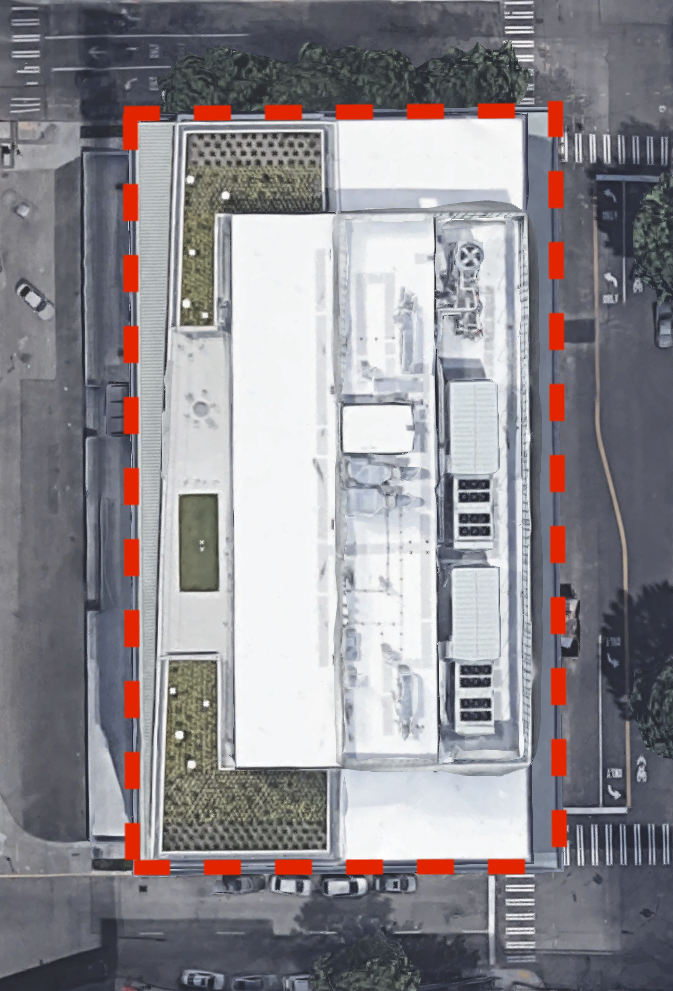}
  \end{overpic}
  \caption*{\small{seattle3}}
\end{subfigure}%
  \caption{Our real world datasets used for evaluation consist of five handheld video sequences of scenes with man-made structures. Here we show sample video frames and the approximate building facade traced on satellite imagery.} 
  \label{fig:datasets}
\end{figure}

In this section, we present our experimental results on both
a synthetic scene, where we know the ground truth results and can hence
quantitatively measure accuracy, as well as some real-world hand-held videos sequences.

\setlength{\fboxsep}{0pt}
\begin{figure*}
\centering %
\begin{subfigure}{0.11\textwidth}
  \vspace{-0.2cm}
  \includegraphics[width=\linewidth]{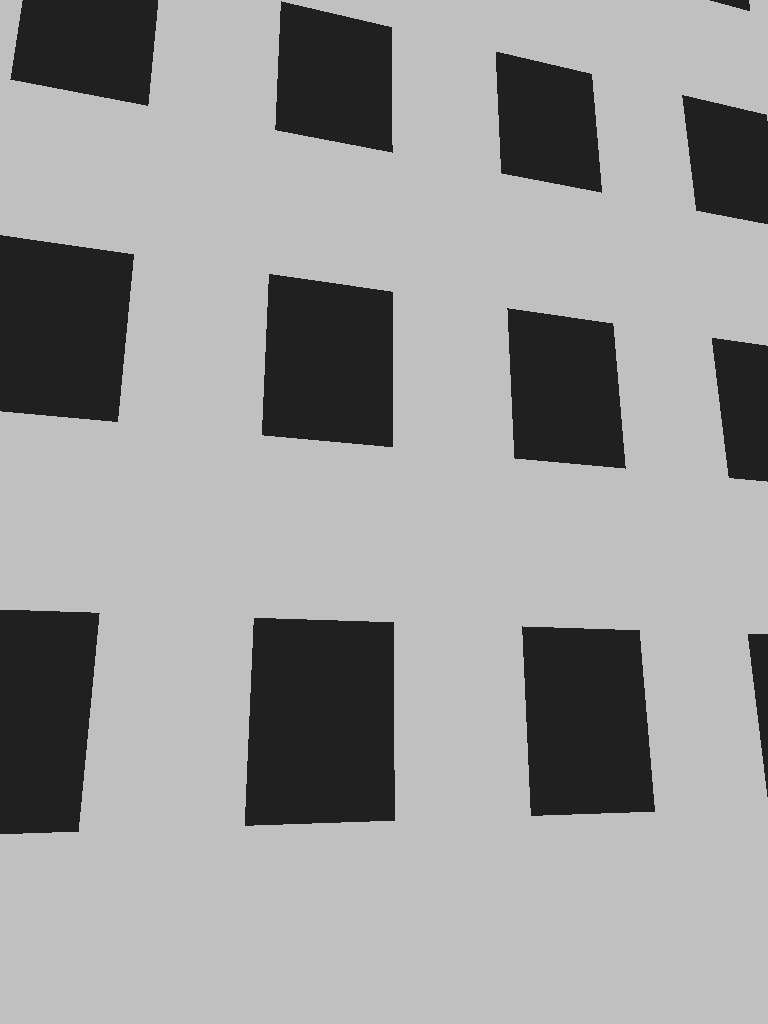}
  \label{subfig:synthetic_sample}
  \vspace{-0.5cm}
  \caption{}
\end{subfigure}\hfil 
  \begin{subfigure}{0.89\textwidth}
\begin{subfigure}{0.25\textwidth}
  \fbox{\includegraphics[width=\linewidth]{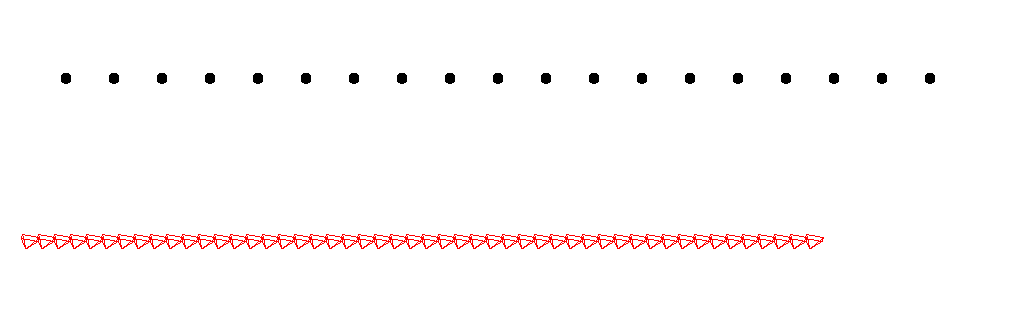}}
\end{subfigure}\hfil 
\begin{subfigure}{0.25\textwidth}
  \fbox{\includegraphics[width=\linewidth]{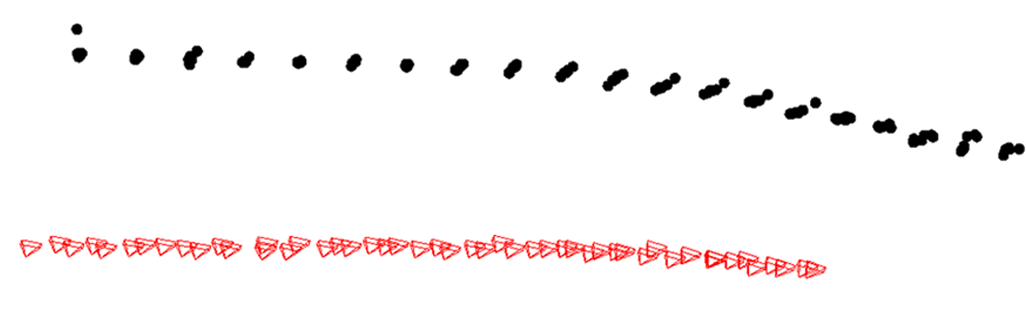}}
\end{subfigure}\hfil 
\begin{subfigure}{0.25\textwidth}
  \fbox{\includegraphics[width=\linewidth]{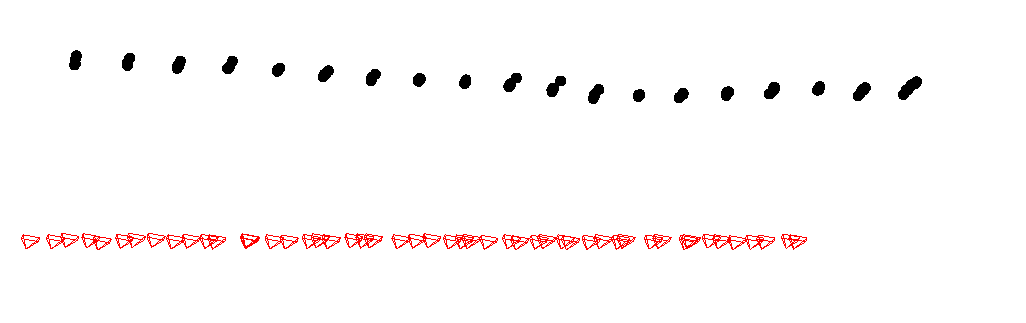}}
\end{subfigure}\hfil 
\begin{subfigure}{0.25\textwidth}
  \fbox{\includegraphics[width=\linewidth]{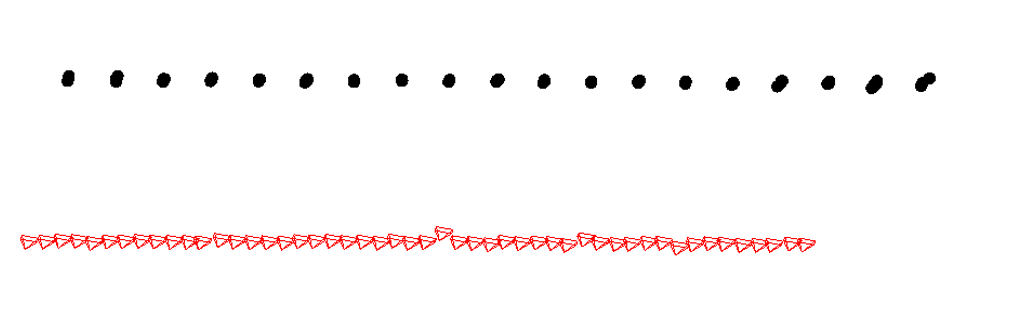}}
\end{subfigure}
\medskip
\begin{subfigure}{0.25\textwidth}
  \fbox{\includegraphics[width=\linewidth]{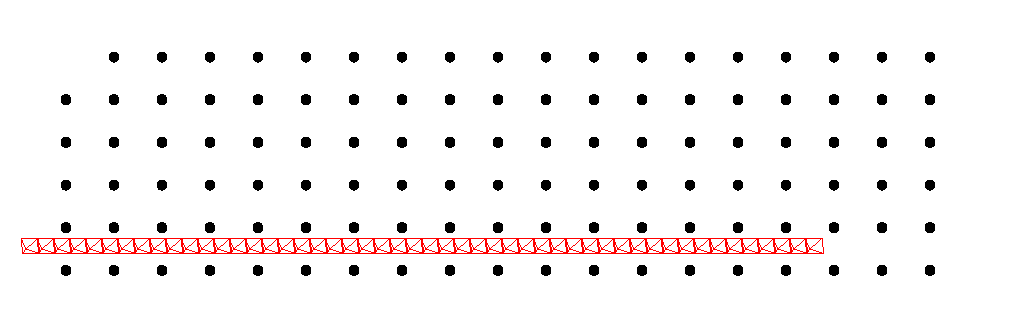}}
  \label{subfig:no_noise}
  \vspace{-0.5cm}
  \caption{}
\end{subfigure}\hfil 
\begin{subfigure}{0.25\textwidth}
  \fbox{\includegraphics[width=\linewidth]{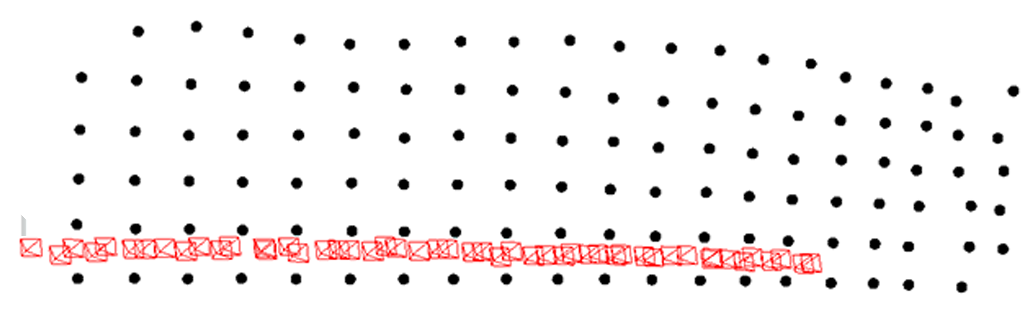}}
  \label{subfig:with_noise}
  \vspace{-0.5cm}
  \caption{}
\end{subfigure}\hfil 
\begin{subfigure}{0.25\textwidth}
  \fbox{\includegraphics[width=\linewidth]{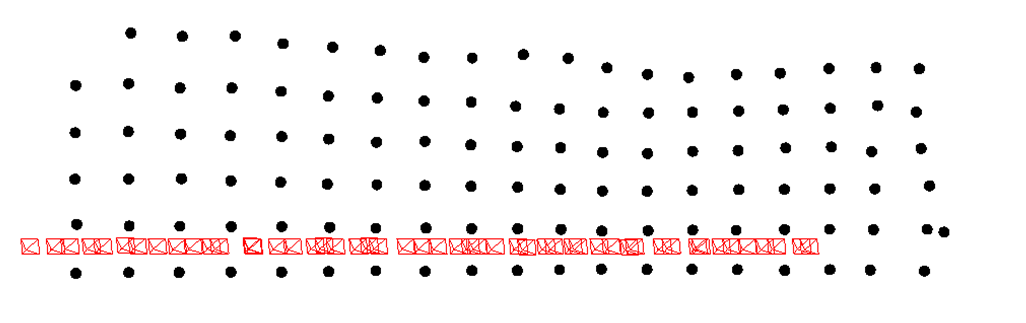}}
  \label{subfig:with_noise_and_vp}
  \vspace{-0.5cm}
  \caption{}
\end{subfigure}\hfil 
\begin{subfigure}{0.25\textwidth}
  \fbox{\includegraphics[width=\linewidth]{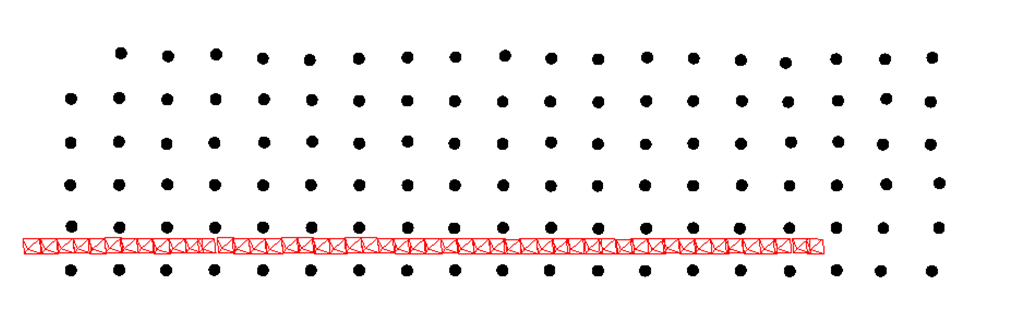}}
  \label{subfig:with_noise_and_vp_and_plane}
  \vspace{-0.5cm}
  \caption{}
\end{subfigure}
  \end{subfigure}
\vspace{-0.5cm}
\caption{Synthetic test case:
{\bf(a)} a sample frame from the sequence. The sequence contains three rows of "windows" on a single plane.
{\bf(b)} Top-down and frontal views of the noise-free reconstruction produced by Theia without added structural constraints: red frusta are shown for the reconstructed camera positions (moving left to right), black points denote the reconstructed 3D window corners.
{\bf(c)} After introducing noise to the 2D point tracks, the reconstruction produced by Theia exhibits both rotation and scale drift.
{\bf(d)} Once vanishing point constraints constraints are added, the rotational drift (bending) is reduced, but scale drift is still present, seen as irregularities in the frame-to-frame camera translations, and non-planarity in the reconstructed points.
{\bf(e)} Once planar constraints are added to the global position estimation, the scale drift is eliminated. \vspace{-0.4cm}
  }
\label{fig:synthetic}
\end{figure*}

\begin{figure}
  \hfill\includegraphics[width=\linewidth]{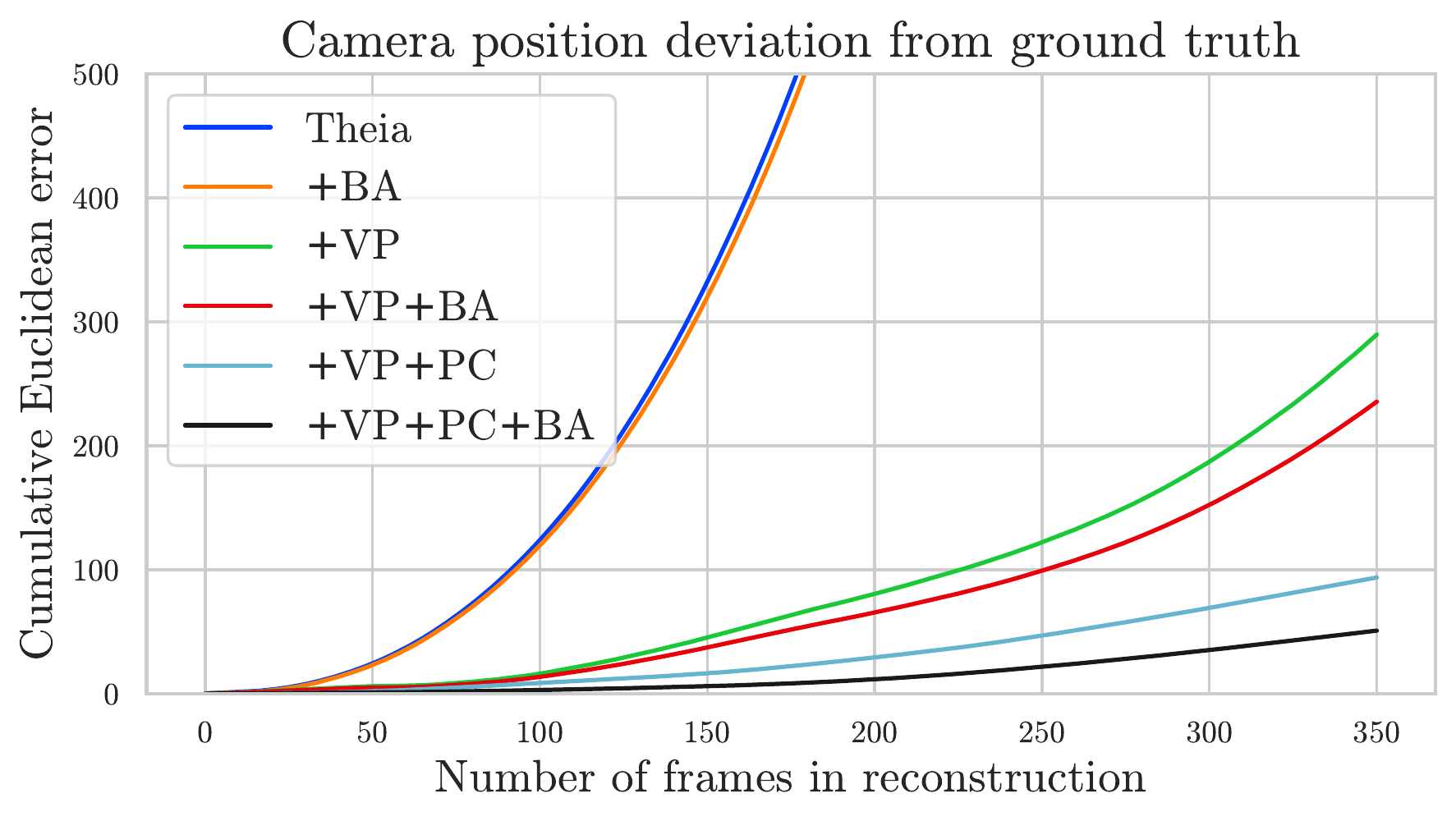}
 \vspace{-0.7cm}
\caption{{\bf Quantitative evaluation}: 
 A comparison of positional drift across variants of Theia with global bundle adjustment (BA), our vanishing point constraints (VP), and our planar constraints (PC). One unit along the vertical axis is equal to the baseline between a pair of cameras. We see that bundle adjustment decreases error across all reconstructions, but cannot fully correct for the positional drift even after the rotational drift has been removed. 
  \vspace{-0.3cm}
}
\label{fig:position_error}
\end{figure}

\subsection{Synthetic scene}

In order to test our algorithm's (and its variants') performance, we constructed
a simple synthetic 3D scene consisting of a three-story building with regularly
spaced windows (Fig.~\ref{fig:synthetic}a). We render this scene from a camera path translating along the length of the building, with small regular fluctuations in elevation. We bypass traditional feature extraction, and instead establish feature tracks by projecting the 3D window corners into each view. We then add synthetic 2D Gaussian noise to these tracks before passing them to the reconstruction algorithm, in order to simulate the correspondence errors which cause drift. Vanishing points are extracted normally, by running our vanishing point estimation on the rendered images. Fig.~\ref{fig:synthetic}c shows that adding Gaussian noise to the 2D point tracks results in accumulated pose error, consisting of both rotational and positional (scale) drift. Fig.~\ref{fig:synthetic}d shows that introducing vanishing point constraints significantly reduces the visible rotational drift (seen as bending in the reconstruction). The remaining errors, caused either by incorrect translation directions or incorrect scale estimates, are shown to be virtually eliminated by introducing global plane constraints (Fig.~\ref{fig:synthetic}e).

In Fig.~\ref{fig:position_error},
we show quantitative results, comparing the positional drift in reconstructions with and without our proposed structural constraints. We measure drift by reconstructing the synthetic sequence shown in Fig.~\ref{fig:synthetic} and comparing the reconstructed poses against ground-truth. For this experiment, we use a sample sequence with 350 frames. Since the reconstructions have scale, rotation, and translation (gauge) ambiguities, we align the reconstructed poses to the ground truth by solving a similarity transformation.

\begin{figure*}
    \centering 
\begin{subfigure}{0.015\textwidth}
  \rotatebox{90}{\texttt{SEATTLE2}}\
\end{subfigure}\hfil 
\begin{subfigure}{0.015\textwidth}
  \rotatebox{90}{\texttt{1167 images}}\
\end{subfigure}\hfil 
\begin{subfigure}{0.1616\textwidth}
  \begin{overpic}[height=\linewidth,width=\linewidth]{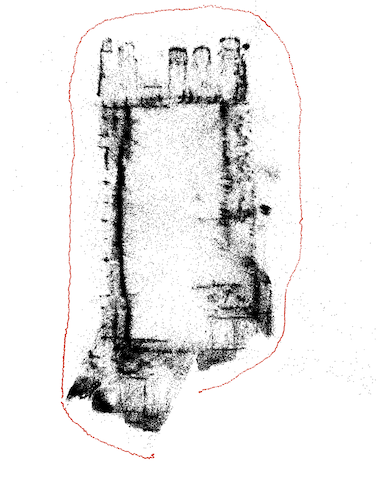}
    \put (35,-7) {\small $423$ s}
  \end{overpic}
\end{subfigure}\hfil 
\begin{subfigure}{0.1616\textwidth}
  \begin{overpic}[height=\linewidth,width=\linewidth]{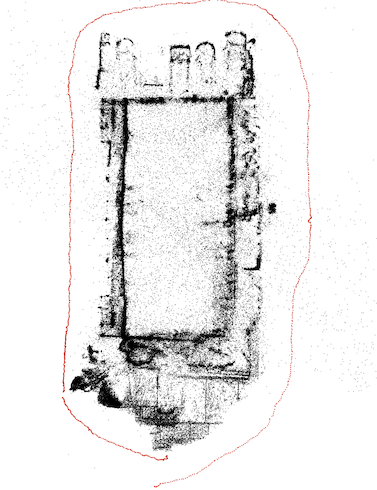}
    \put (35,-7) {\small $1282$ s}
  \end{overpic}
\end{subfigure}\hfil 
\begin{subfigure}{0.1616\textwidth}
  \begin{overpic}[height=\linewidth,width=\linewidth]{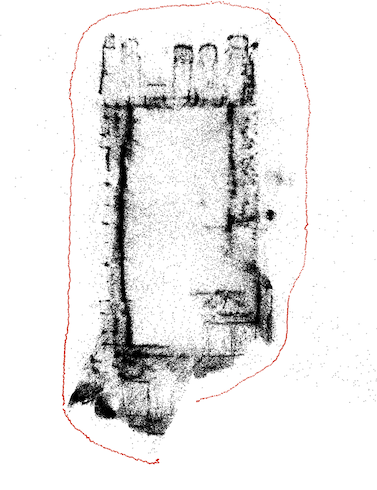}
    \put (35,-7) {\small $416$ s}
  \end{overpic}
\end{subfigure}\hfil 
\begin{subfigure}{0.1616\textwidth}
  \begin{overpic}[height=\linewidth,width=\linewidth]{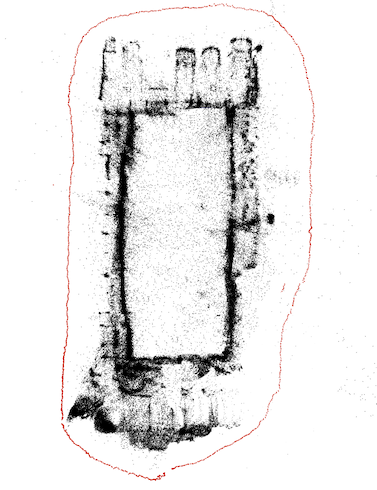}
    \put (35,-7) {\small $475$ s}
  \end{overpic}
\end{subfigure}\hfil 
\begin{subfigure}{0.1616\textwidth}
  \begin{overpic}[height=\linewidth,width=\linewidth]{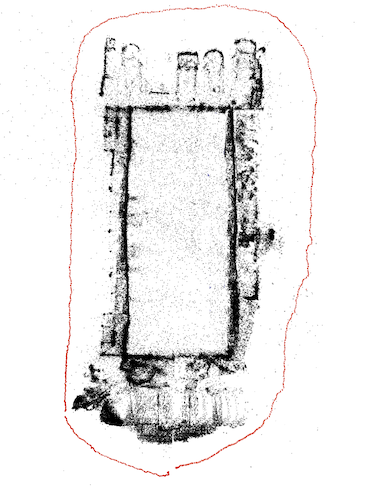}
    \put (35,-7) {\small $607$ s}
  \end{overpic}
\end{subfigure}\hfil 
\begin{subfigure}{0.1616\textwidth}
  \begin{overpic}[height=\linewidth,width=\linewidth]{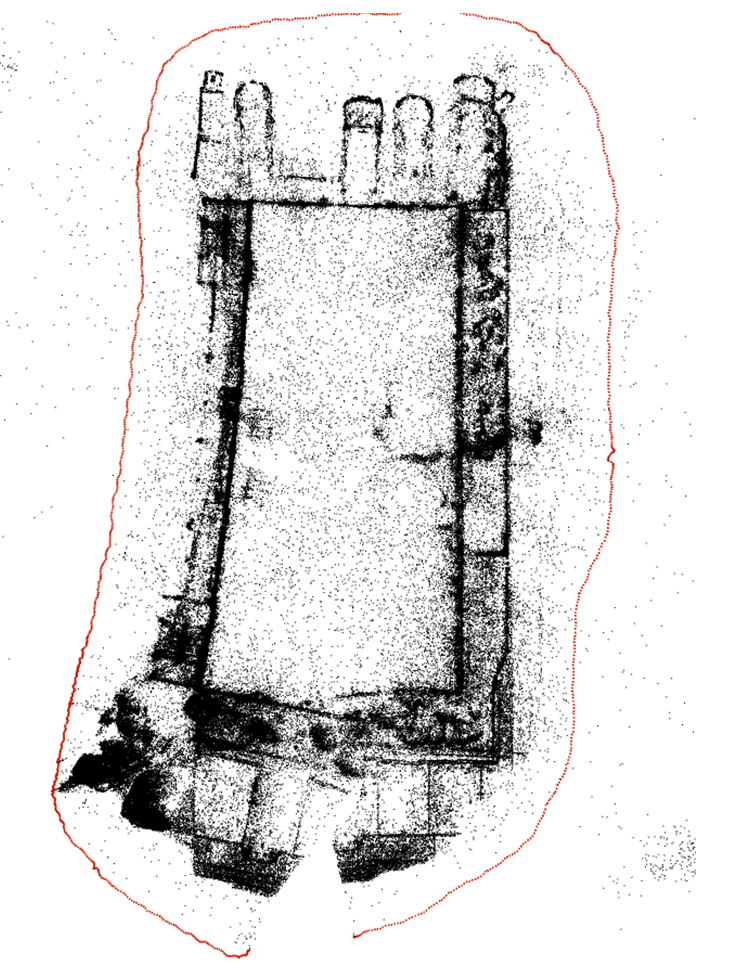}
    \put (25,-7) {\small $13498$ s}
  \end{overpic}
\end{subfigure}

\medskip
\begin{subfigure}{0.015\textwidth}
  \rotatebox{90}{\texttt{MORE\char`_HALF}}\
\end{subfigure}\hfil 
\begin{subfigure}{0.015\textwidth}
  \rotatebox{90}{\texttt{1380 images}}\
\end{subfigure}\hfil 
\begin{subfigure}{0.1616\textwidth}
  \begin{overpic}[width=\linewidth]{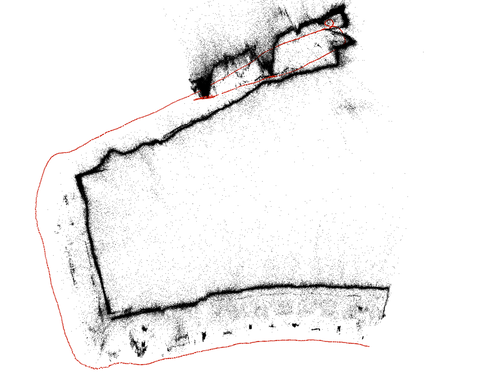}
    \put (50,-5) {\small 263 s}
  \end{overpic}
\end{subfigure}\hfil 
\begin{subfigure}{0.1616\textwidth}
  \begin{overpic}[width=\linewidth]{figures/images/more_half/vanilla_bundle.png}
    \put (50,-5) {\small 857 s}
  \end{overpic}
\end{subfigure}\hfil 
\begin{subfigure}{0.1616\textwidth}
  \begin{overpic}[width=\linewidth]{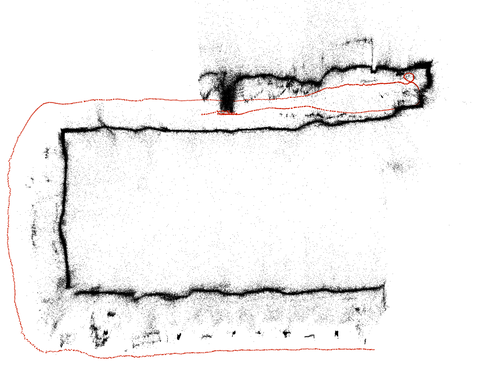}
    \put (50,-5) {\small 268 s}
  \end{overpic}
\end{subfigure}\hfil 
\begin{subfigure}{0.1616\textwidth}
  \begin{overpic}[width=\linewidth]{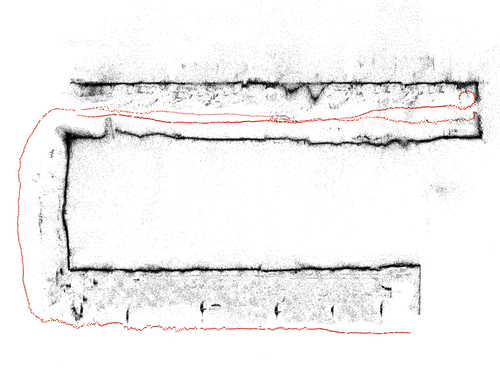}
    \put (50,-5) {\small 364 s}
  \end{overpic}
\end{subfigure}\hfil 
\begin{subfigure}{0.1616\textwidth}
  \begin{overpic}[width=\linewidth]{figures/images/more_half/plane_bundle.png}
    \put (50,-5) {\small 464 s}
  \end{overpic}
\end{subfigure}\hfil 
\begin{subfigure}{0.1616\textwidth}
  \begin{overpic}[width=\linewidth]{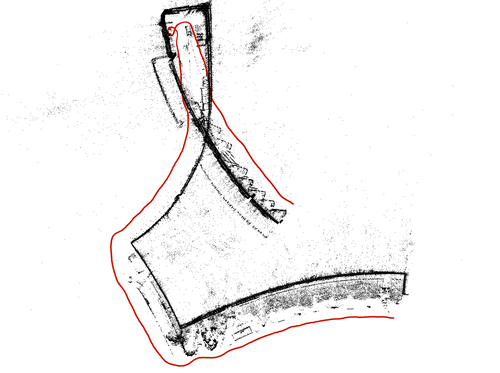}
    \put (50,-5) {\small 12579 s}
  \end{overpic}
\end{subfigure}

\medskip
\begin{subfigure}{0.015\textwidth}
  \rotatebox{90}{\texttt{SEATTLE1}}\
\end{subfigure}\hfil 
\begin{subfigure}{0.015\textwidth}
  \rotatebox{90}{\texttt{892 images}}\
\end{subfigure}\hfil 
\begin{subfigure}{0.1616\textwidth}
  \begin{overpic}[height=\linewidth,width=\linewidth]{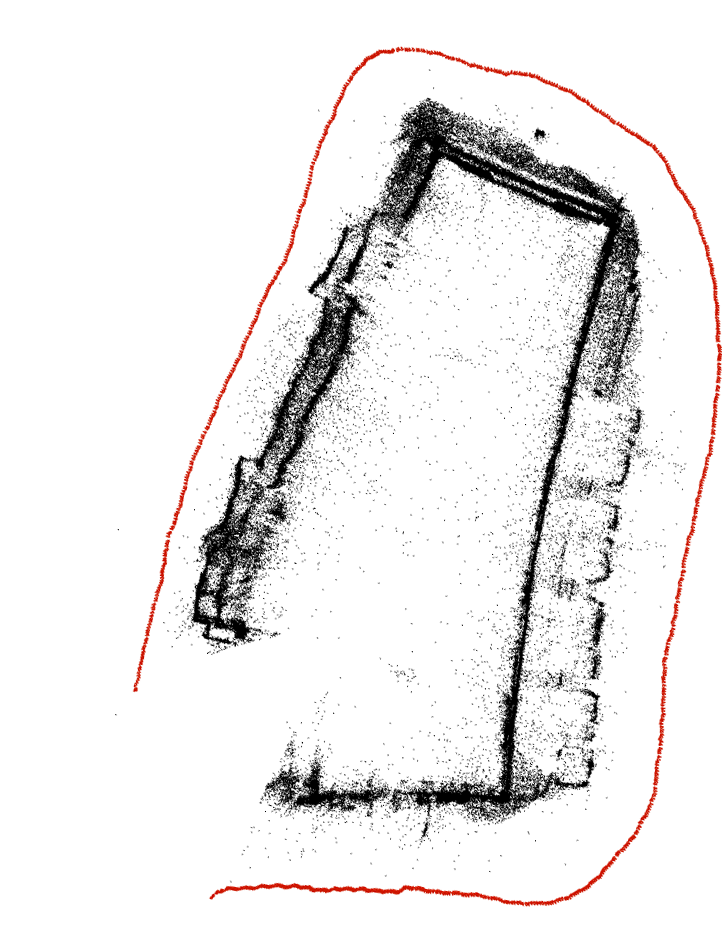}
    \put (50,-5) {\small $111$ s}
  \end{overpic}
\end{subfigure}\hfil 
\begin{subfigure}{0.1616\textwidth}
  \begin{overpic}[height=\linewidth,width=\linewidth]{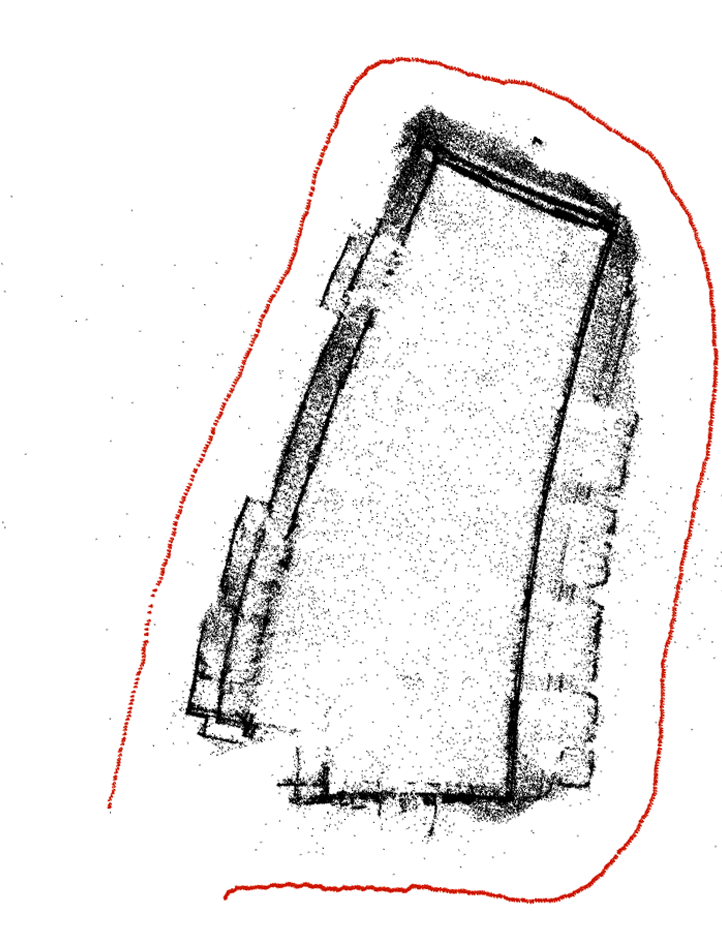}
    \put (50,-5) {\small $200$ s}
  \end{overpic}
\end{subfigure}\hfil 
\begin{subfigure}{0.1616\textwidth}
  \begin{overpic}[height=\linewidth,width=\linewidth]{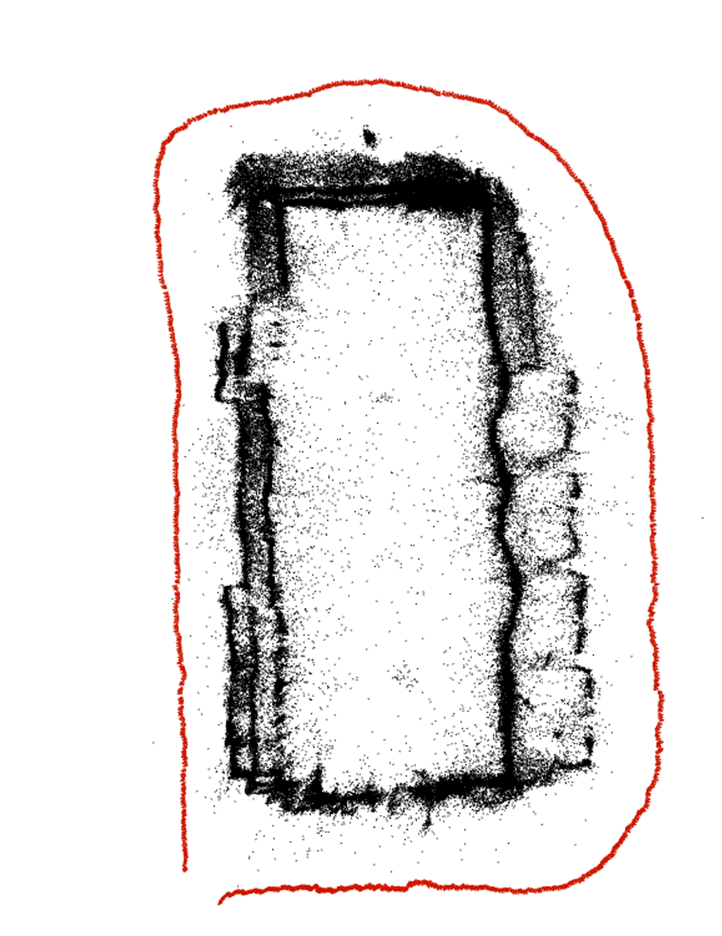}
    \put (50,-5) {\small $107$ s}
  \end{overpic}
\end{subfigure}\hfil 
\begin{subfigure}{0.1616\textwidth}
  \begin{overpic}[height=\linewidth,width=\linewidth]{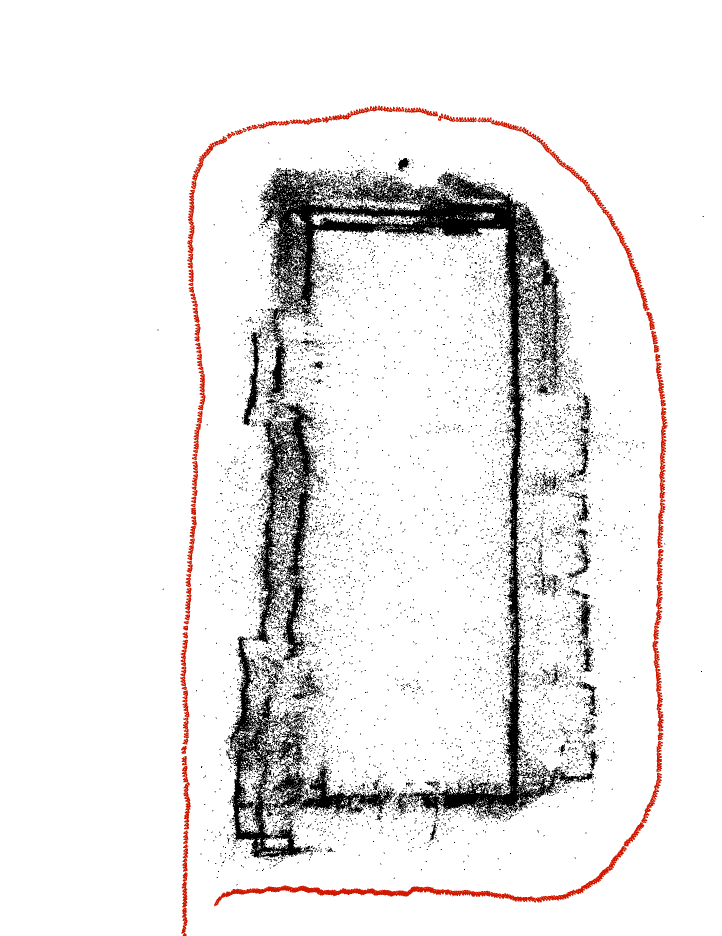}
    \put (50,-5) {\small $132$ s}
  \end{overpic}
\end{subfigure}\hfil 
\begin{subfigure}{0.1616\textwidth}
  \begin{overpic}[height=\linewidth,width=\linewidth]{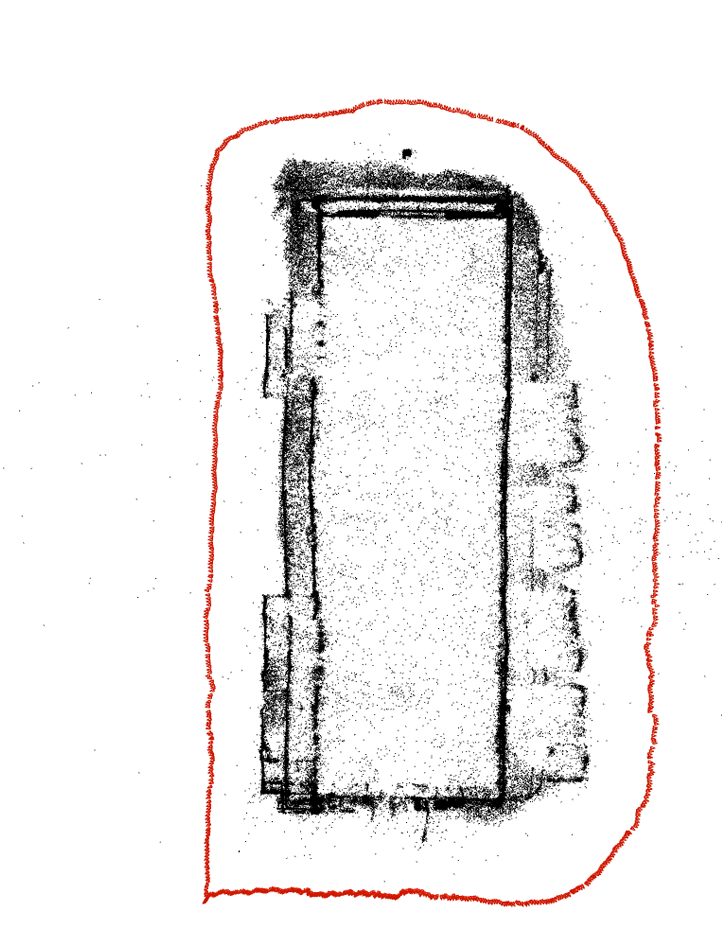}
    \put (50,-5) {\small $194$ s}
  \end{overpic}
\end{subfigure}\hfil 
\begin{subfigure}{0.1616\textwidth}
  \begin{overpic}[height=\linewidth,width=\linewidth]{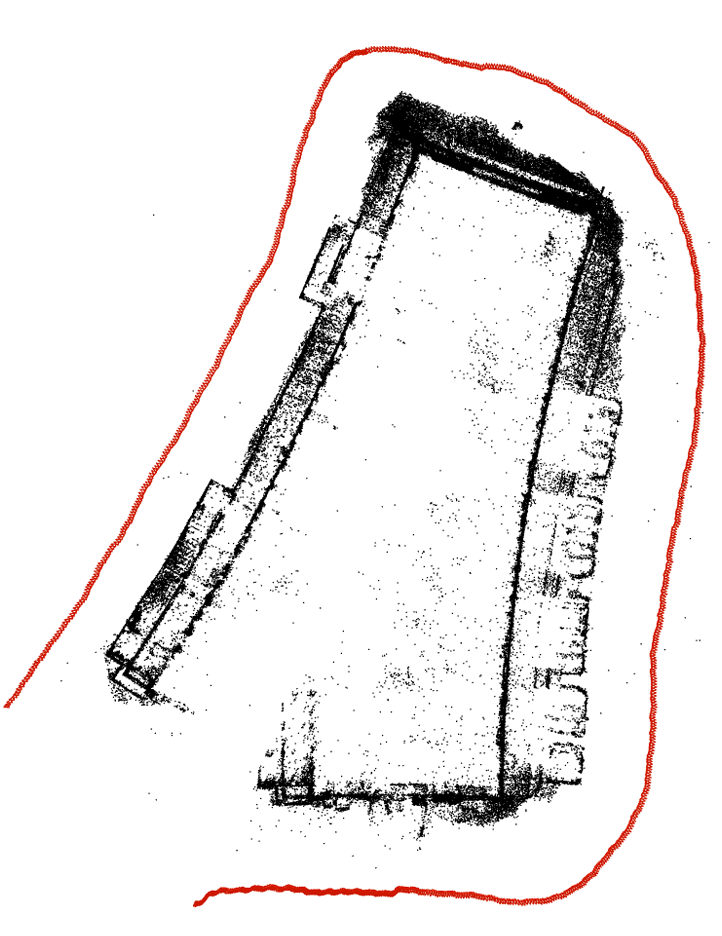}
    \put (50,-5) {\small $1368$ s}
  \end{overpic}
\end{subfigure}

\medskip
\begin{subfigure}{0.015\textwidth}
  \rotatebox{90}{\texttt{ATLANTA1}}\
\end{subfigure}\hfil 
\begin{subfigure}{0.015\textwidth}
  \rotatebox{90}{\texttt{722 images}}\
\end{subfigure}\hfil 
\begin{subfigure}{0.1616\textwidth}
\begin{center}
  \begin{overpic}[height=\linewidth]{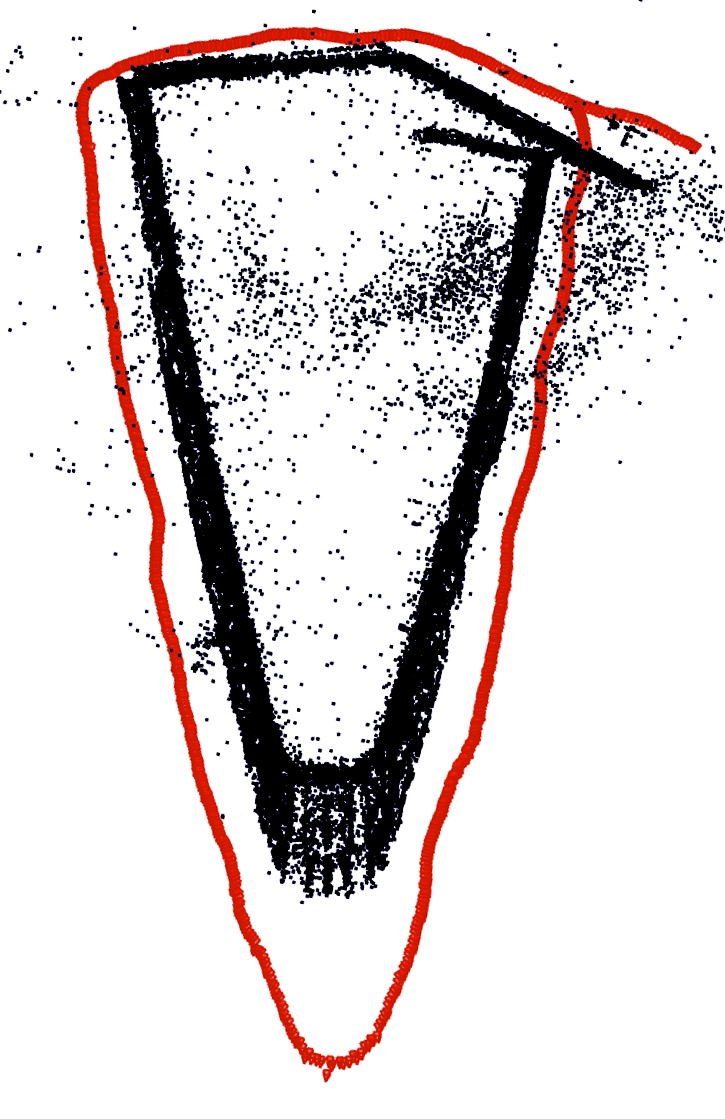}
    \put (35,-5) {\small $89$ s}
  \end{overpic}
   \end{center}
\end{subfigure}\hfil 
\begin{subfigure}{0.1616\textwidth}
\begin{center}
  \begin{overpic}[height=\linewidth]{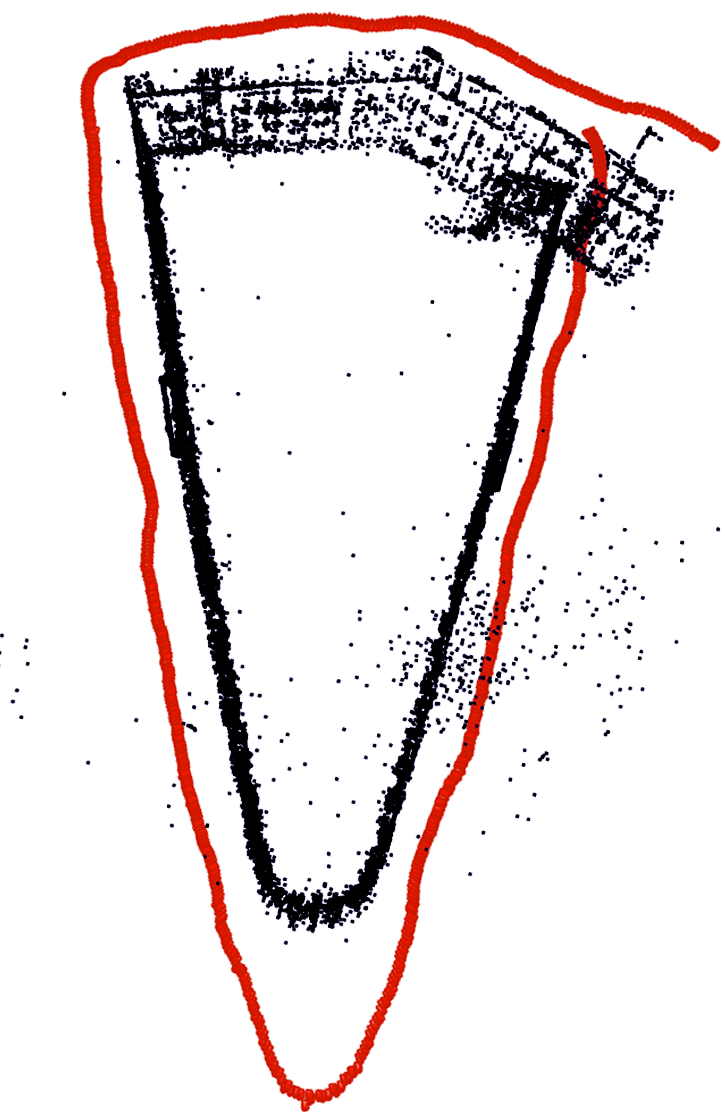}
    \put (35,-5) {\small $96$ s}
  \end{overpic}
   \end{center}
\end{subfigure}\hfil 
\begin{subfigure}{0.1616\textwidth}
\begin{center}
  \begin{overpic}[height=\linewidth]{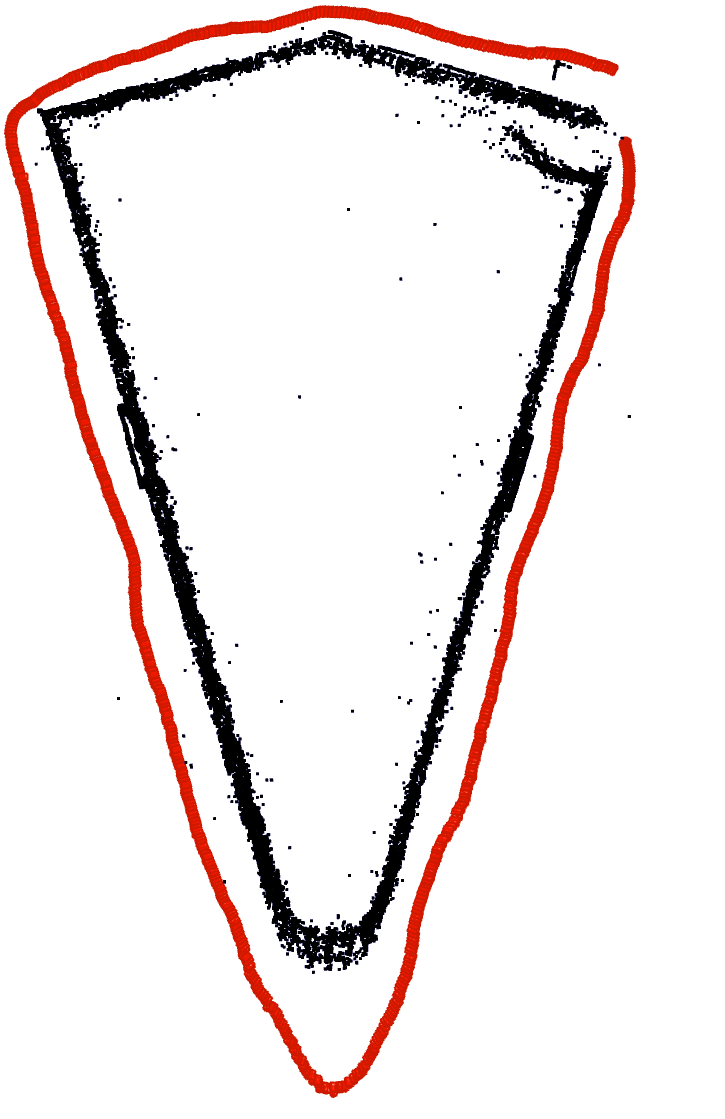}
    \put (35,-5) {\small $90$ s}
  \end{overpic}
  \end{center}
\end{subfigure}\hfil 
\begin{subfigure}{0.1616\textwidth}
\begin{center}
  \begin{overpic}[height=\linewidth]{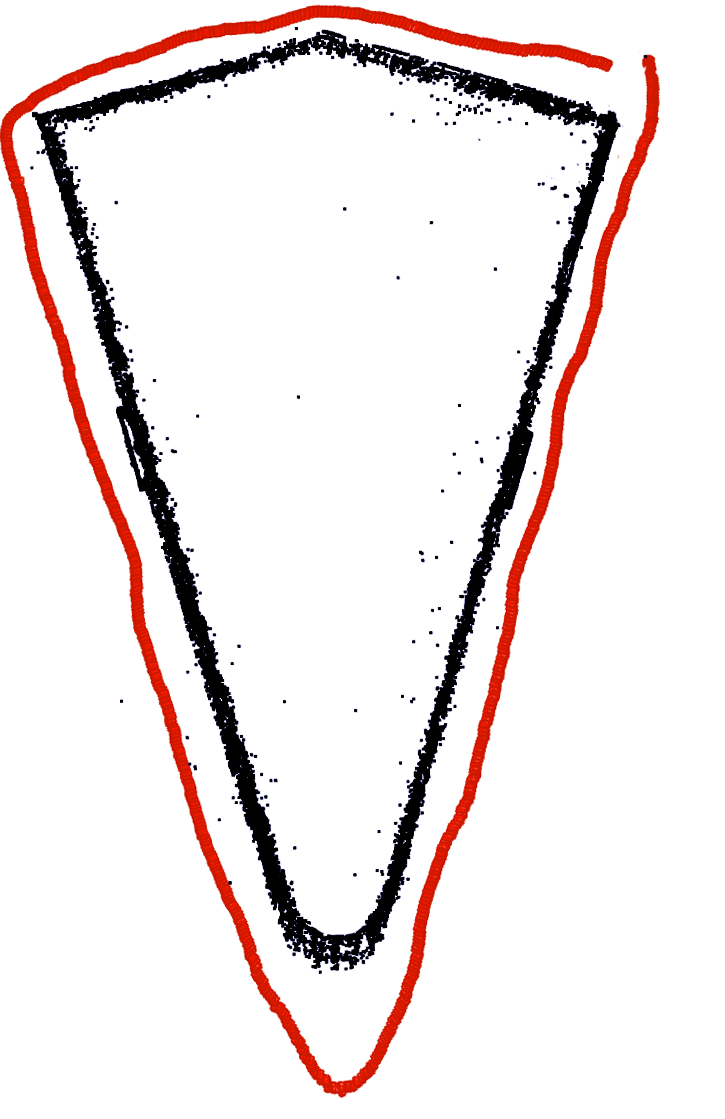}
    \put (35,-5) {\small $98$ s}
  \end{overpic}
 \end{center}
\end{subfigure}\hfil 
\begin{subfigure}{0.1616\textwidth}
\begin{center}
  \begin{overpic}[height=\linewidth]{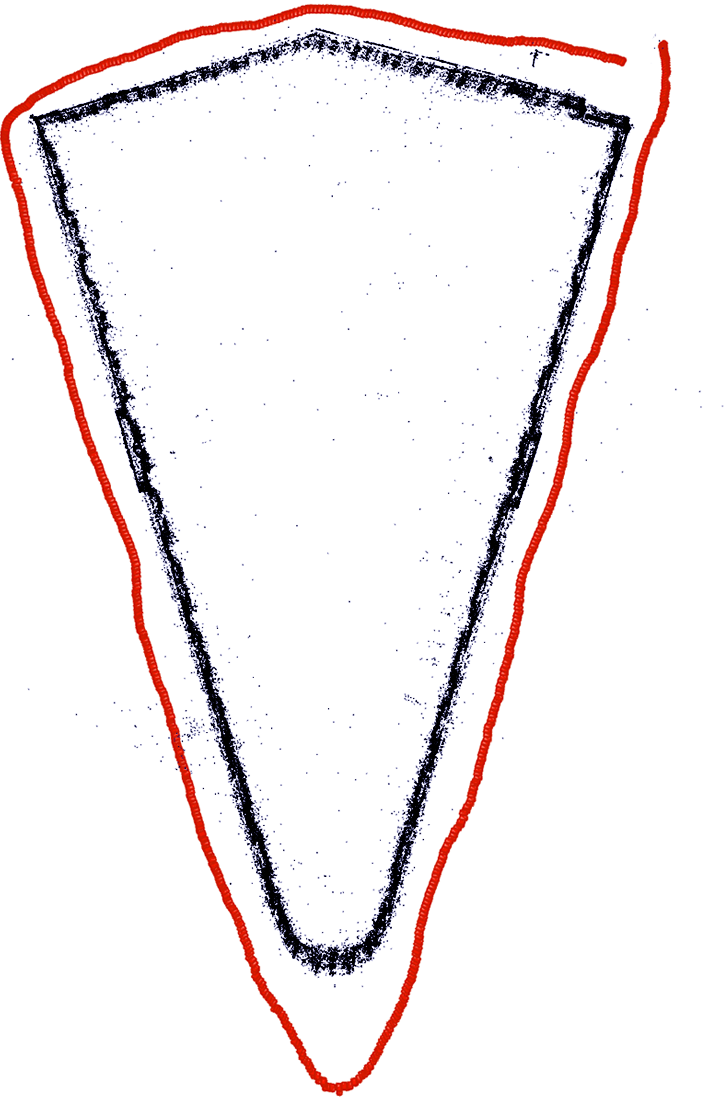}
    \put (35,-5) {\small $99$ s}
  \end{overpic}
   \end{center}
\end{subfigure}\hfil 
\begin{subfigure}{0.1616\textwidth}
\begin{center}
  \begin{overpic}[height=\linewidth]{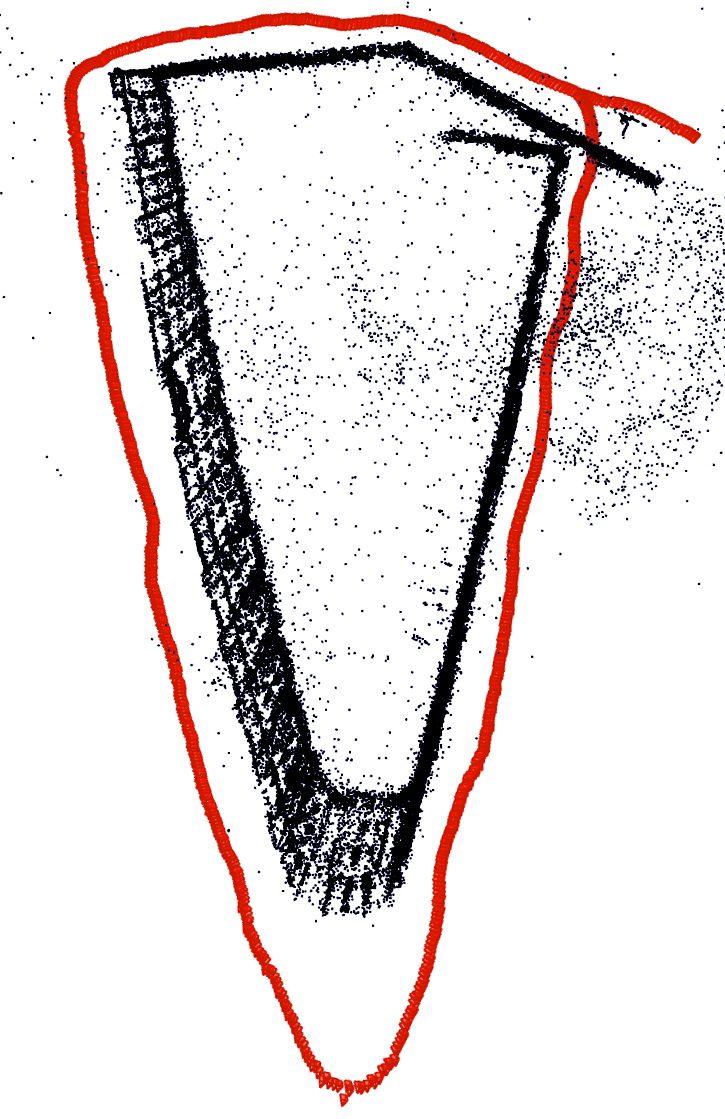}
    \put (35,-5) {\small $1589$ s}
  \end{overpic}
   \end{center}
\end{subfigure}

\medskip
\begin{subfigure}{0.015\textwidth}
  \rotatebox{90}{\texttt{SEATTLE3}}\
\end{subfigure}\hfil 
\begin{subfigure}{0.015\textwidth}
  \rotatebox{90}{\texttt{1024 images}}\
\end{subfigure}\hfil 
\begin{subfigure}{0.1616\textwidth}
\begin{center}
  \begin{overpic}[height=\linewidth]{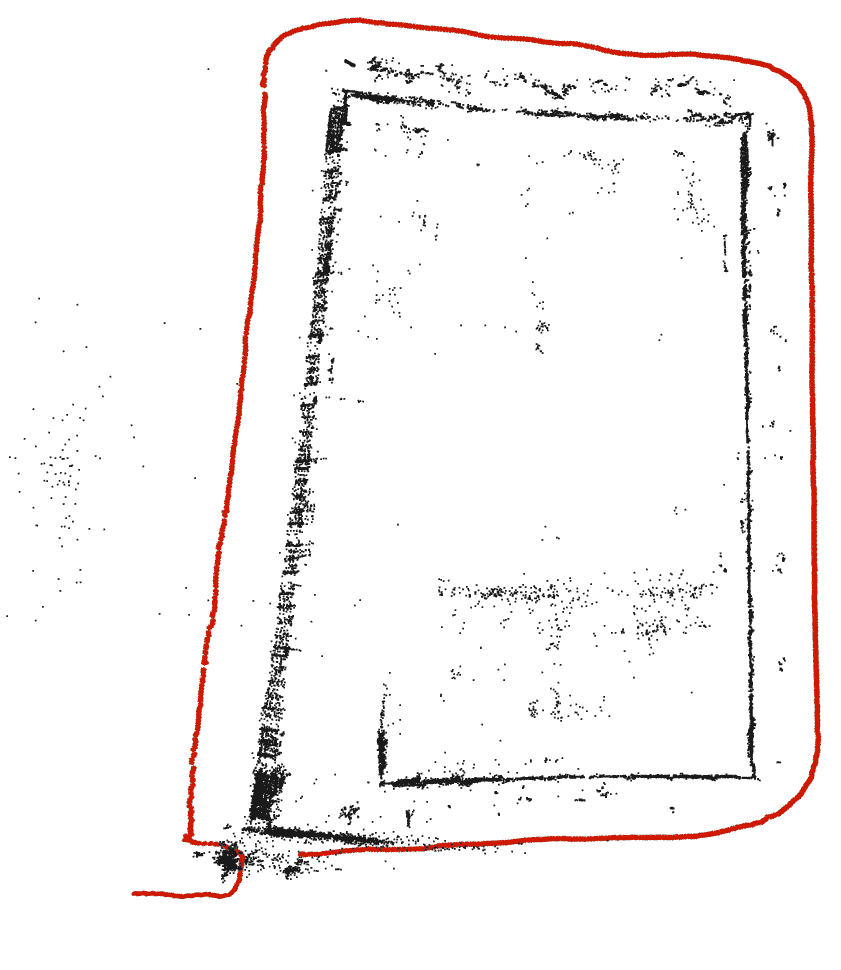}
    \put (40,-5) {\small $198$ s}
  \end{overpic}
  \end{center}
  \captionsetup{labelfont=bf}
  \caption{\\Theia}
\end{subfigure}\hfil 
\begin{subfigure}{0.1616\textwidth}\begin{center}
  \begin{overpic}[height=\linewidth]{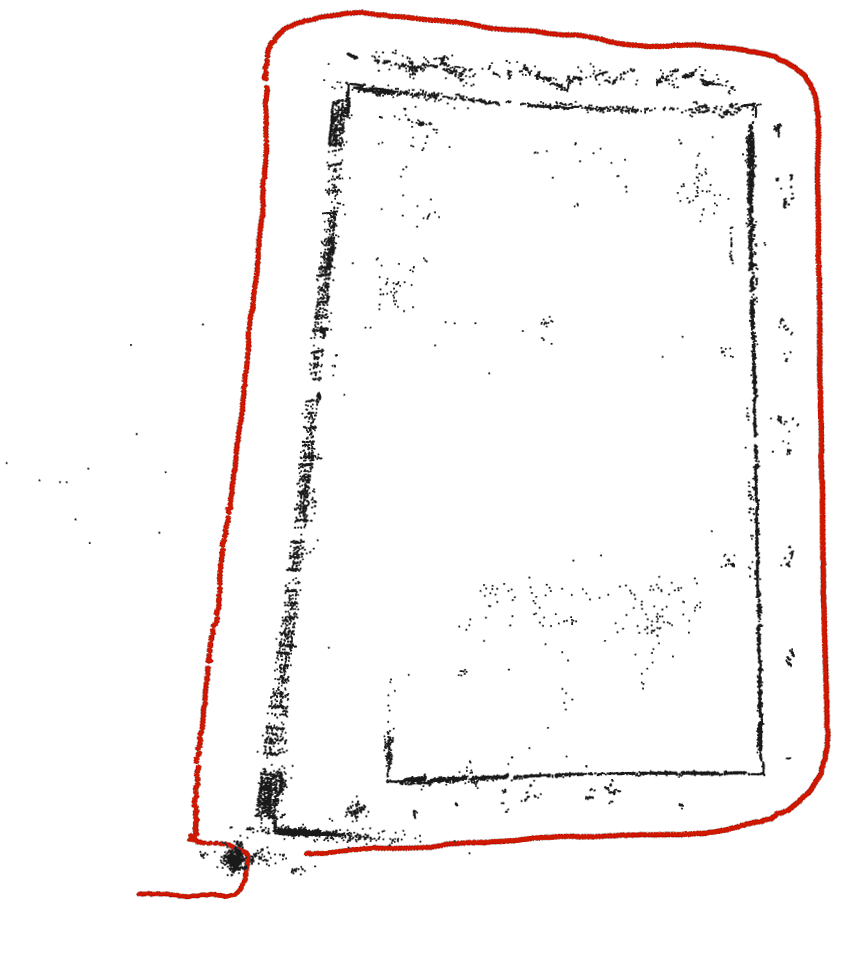}
    \put (40,-5) {\small $288$ s}
  \end{overpic}\end{center}
  \captionsetup{labelfont=bf}
  \caption{\\Theia+BA}
\end{subfigure}\hfil 
\begin{subfigure}{0.1616\textwidth}\begin{center}
  \begin{overpic}[height=\linewidth]{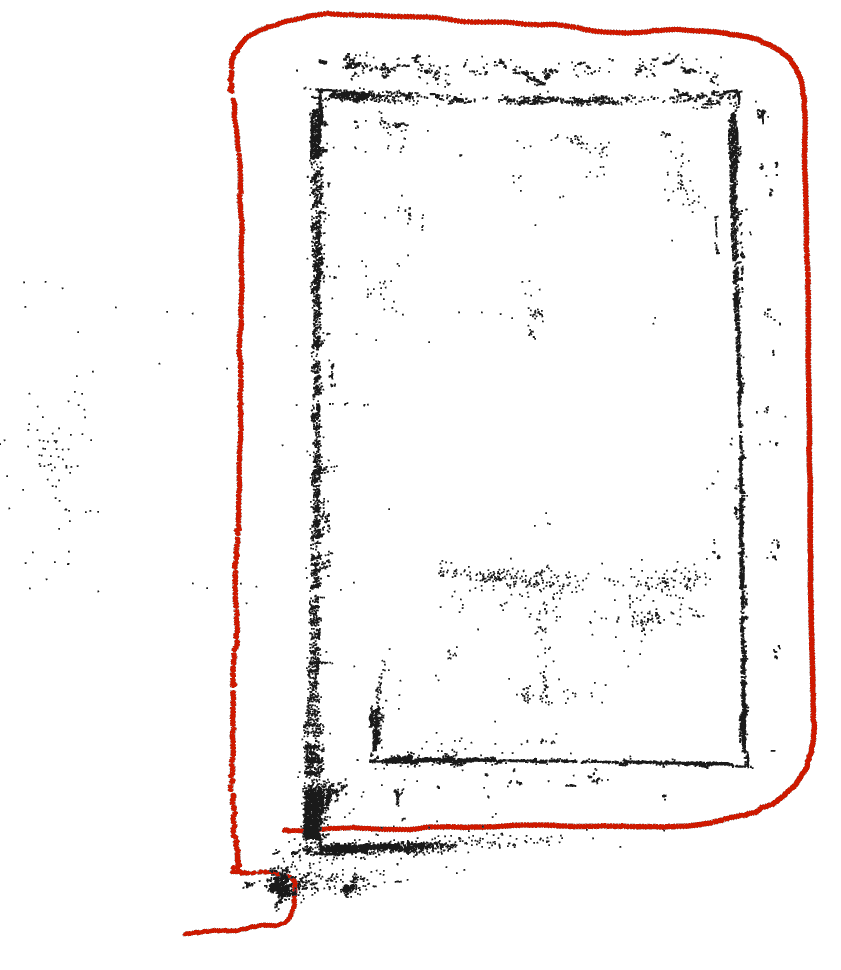}
    \put (40,-5) {\small $195$ s}
  \end{overpic}\end{center}
  \captionsetup{labelfont=bf}
  \caption{\\VP Only}
\end{subfigure}\hfil 
\begin{subfigure}{0.1616\textwidth}\begin{center}
  \begin{overpic}[height=\linewidth]{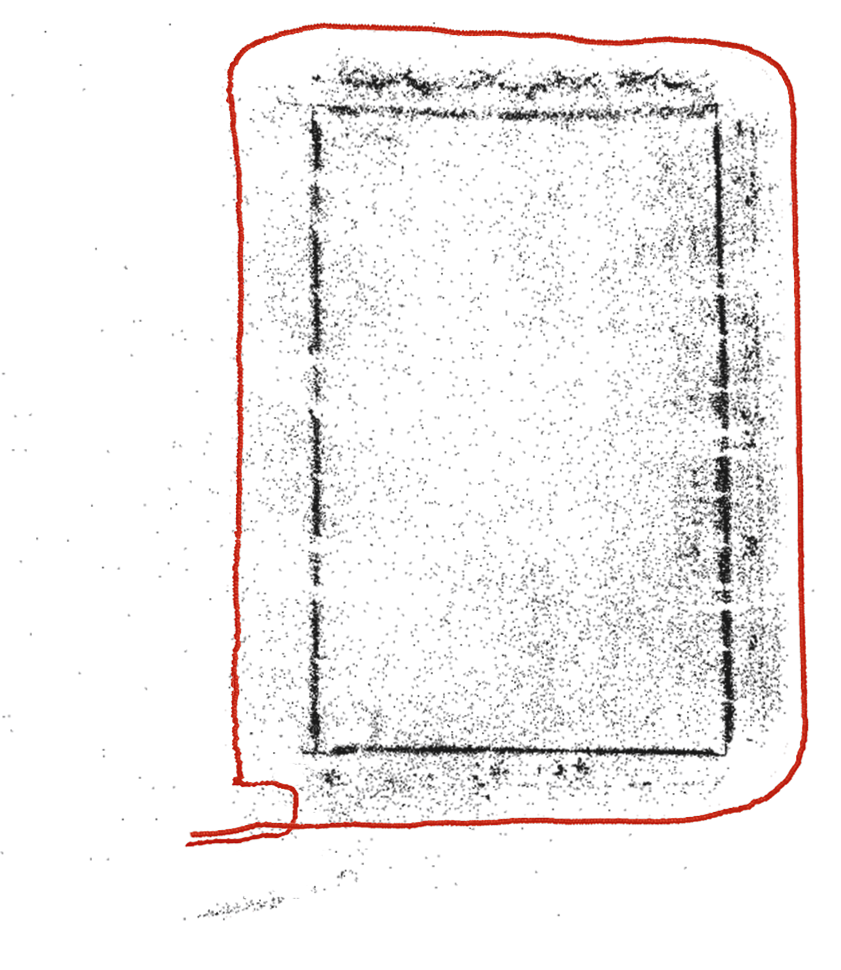}
    \put (40,-5) {\small $208$ s}
  \end{overpic}\end{center}
  \captionsetup{labelfont=bf}
  \caption{\\Ours}
\end{subfigure}\hfil 
\begin{subfigure}{0.1616\textwidth}\begin{center}
  \begin{overpic}[height=\linewidth]{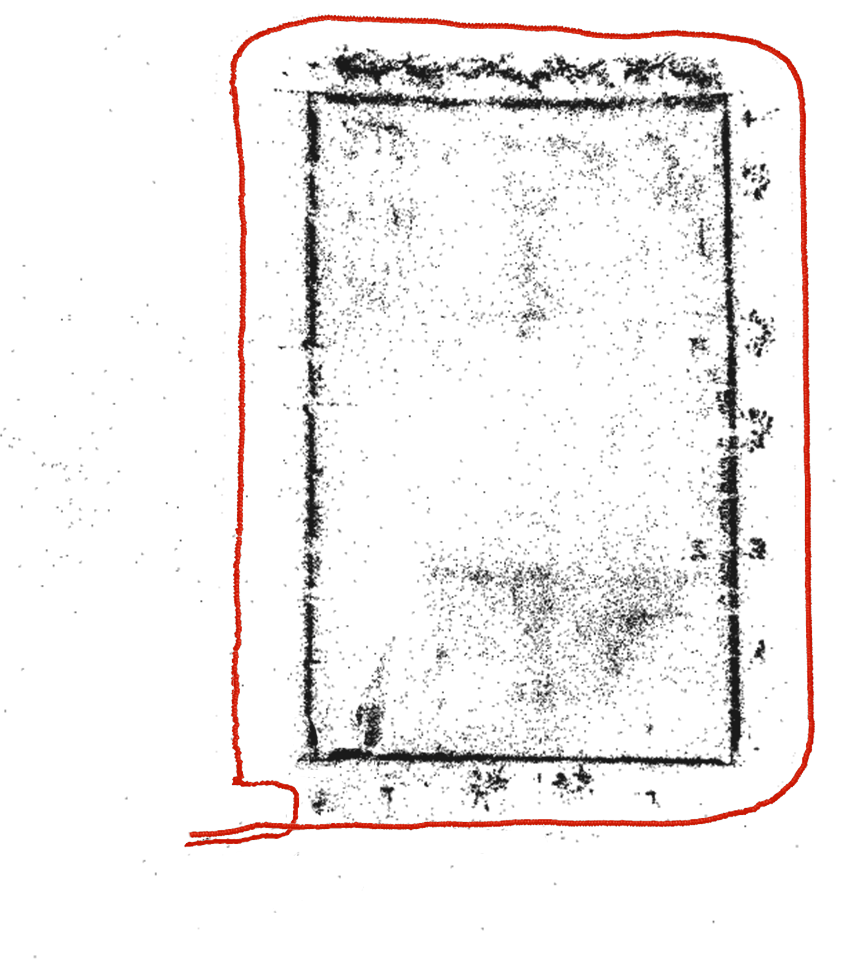}
    \put (40,-5) {\small $251$ s}
  \end{overpic}\end{center}
  \captionsetup{labelfont=bf}
  \caption{\\Ours+BA}
\end{subfigure}\hfil 
\begin{subfigure}{0.1616\textwidth}\begin{center}
  \begin{overpic}[height=\linewidth]{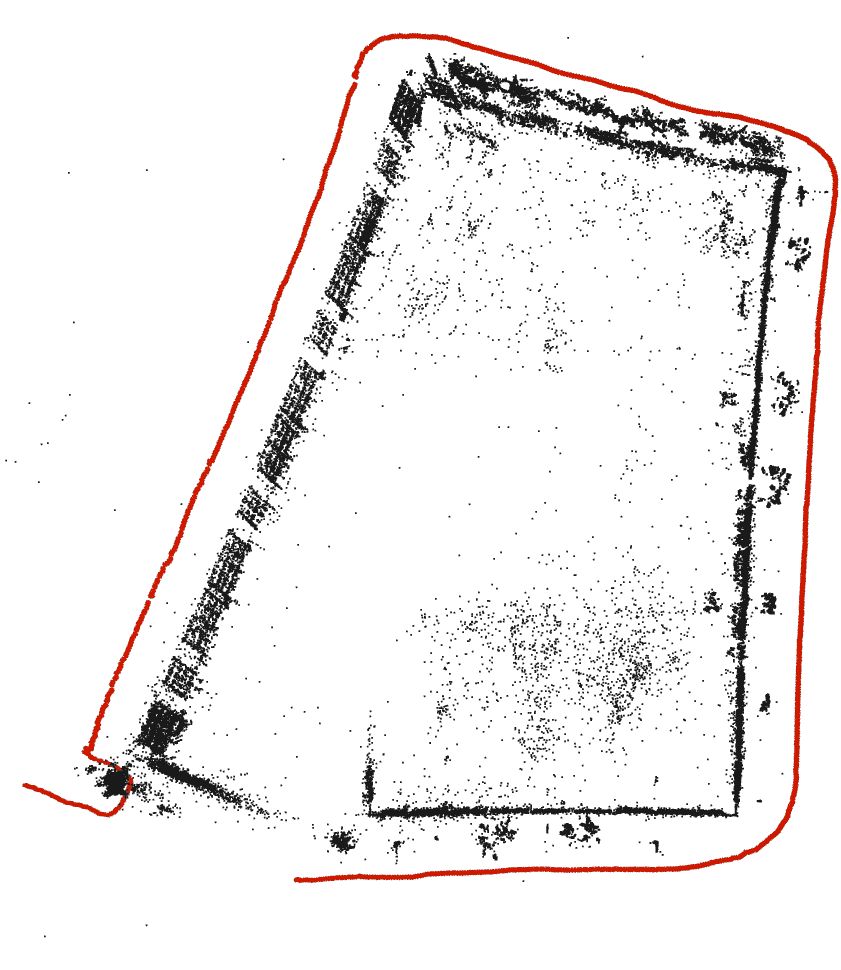}
    \put (40,-5) {\small $2514$ s}
  \end{overpic}\end{center}
  \captionsetup{labelfont=bf}
  \caption{\\COLMAP}
\end{subfigure}

\caption{{\bf Qualitative comparison:} top-down views of the real-world datasets using different reconstruction methods: {\bf (a)} Theia without added structural constraints or global bundle adjustment (BA), {\bf (b)} Theia without added constraints but with BA, {\bf (c)} Theia with added vanishing point constraints (VP), but no planar constraints (PC) or BA {\bf (d)} Theia with both VP and PC, but without BA {\bf (e)} Theia with VP, PC, and BA, {\bf (f)} COLMAP, with BA.
None of the reconstructions use any form of explicit loop closure; all examples use window-based feature matching with a window size of 100 frames. See supplemental for a discussion about loop closure. Inset values indicate reconstruction time, excluding feature matching, measured on a MacBook Pro with a 6-core 2.9 GHz Intel processor and 32 GB of memory. \vspace{-0.7cm}}
\label{fig:qualitative}
\end{figure*}

\subsection{Real-world videos}
\label{sec:real_world} 
Typical SLAM and SfM benchmarks use static (or global shutter) cameras with wide fields-of-view and contain widely varying viewpoints of the scene. These configurations are chosen because they are the least susceptible to drift --- tracks will be longer, feature positions more precise, and triangulation angles wider, thus reducing the effects of low-frequency pose error. In this paper, we focus instead on the converse: configurations which are \emph{most} susceptible, i.e. low-field of view, rolling-shutter, handheld videos, where objects are only seen from a small range of viewpoints. These types of captures also more accurately reflect the type of sequence that might be captured by a layperson with a handheld camera-phone.

With this in mind, we captured five handheld video sequences of man-made environments (Fig.~\ref{fig:datasets}). We used smartphone cameras in portrait orientation, and mostly captured sequences walking along building facades. These sequences are intended to showcase difficult configurations which typically result in drift, since point tracks do not persist for many frames, and therefore cannot establish long-term constraints.
In Fig.~\ref{fig:qualitative}, we show a comparison to our baseline method, as well as to COLMAP \cite{Schonberger2016}, a popular incremental SfM system. We can see that the baseline methods (Theia and COLMAP) both exhibit significant drift, and the addition of our constraints to Theia results in a system that produces drift-free reconstructions very quickly (a fraction of the time needed for COLMAP). It is important to note that none of these results use any form of explicit loop closure. This is intended to more visibly demonstrate the effects of drift, as automatic loop closure is not always an option (for sequences without enough loop overlap) and, when it is, does not always remove the effects of drift. It is also possible for automatic loop closure to introduce other sources of error, such as false matches between distant views, especially in sequences like ours, which contain significant repetitive structure. In the supplemental material, we show experiments with automatic loop closure enabled. 

\renewcommand{\tabcolsep}{1pt}
\begin{table}[]
\small
\begin{tabular}{rcccccccc}
\multicolumn{2}{c}{SEATTLE1}                    & \multicolumn{2}{c}{ SEATTLE2} & \multicolumn{2}{c}{SEATTLE3}                    & \multicolumn{2}{c}{ATLANTA1}                    \\ \cline{1-8}
\multicolumn{1}{|c|}{Pos.} & \multicolumn{1}{c|}{Orient.} & \multicolumn{1}{c|}{Pos.} & \multicolumn{1}{c|}{Orient.} & \multicolumn{1}{c|}{Pos.} & \multicolumn{1}{c|}{Orient.} & \multicolumn{1}{c|}{Pos.} & \multicolumn{1}{c|}{Orient.}\\ \cline{1-8} 
\multicolumn{1}{|c|}{73.51} & \multicolumn{1}{c|}{12.63\degree} & \multicolumn{1}{c|}{45.04} & \multicolumn{1}{c|}{8.70\degree}  &
\multicolumn{1}{|c|}{48.06} & \multicolumn{1}{c|}{10.23\degree} & \multicolumn{1}{c|}{36.22} & \multicolumn{1}{c|}{15.71\degree}  & \multicolumn{1}{l}{\small Theia} \\ \cline{1-8}
\multicolumn{1}{|c|}{30.01} & \multicolumn{1}{c|}{$12.60\degree$} & \multicolumn{1}{c|}{11.63} & \multicolumn{1}{c|}{8.56\degree}  &
\multicolumn{1}{|c|}{46.85} & \multicolumn{1}{c|}{9.19\degree} & \multicolumn{1}{c|}{32.02} & \multicolumn{1}{c|}{14.26\degree}  &\multicolumn{1}{l}{\small Theia+BA} \\ \cline{1-8} 
\multicolumn{1}{|c|}{16.98} & \multicolumn{1}{c|}{\bf{1.72\degree}} & \multicolumn{1}{c|}{25.00} & \multicolumn{1}{c|}{\bf{2.01\degree}}  &
\multicolumn{1}{|c|}{40.07} & \multicolumn{1}{c|}{2.70\degree} & \multicolumn{1}{c|}{18.11} & \multicolumn{1}{c|}{2.18\degree}  &\multicolumn{1}{l}{\small Theia+VP} \\ \cline{1-8} 
\multicolumn{1}{|c|}{7.42} & \multicolumn{1}{c|}{\bf{1.72\degree}} & \multicolumn{1}{c|}{3.61} & \multicolumn{1}{c|}{\bf{2.01\degree}}  &
\multicolumn{1}{|c|}{4.22} & \multicolumn{1}{c|}{2.70\degree} & \multicolumn{1}{c|}{6.50} & \multicolumn{1}{c|}{2.18\degree}  &\multicolumn{1}{l}{\small Ours} \\ \cline{1-8} 
\multicolumn{1}{|c|}{\bf{1.91}} & \multicolumn{1}{c|}{\bf{1.72\degree}} & \multicolumn{1}{c|}{\bf{3.34}} & \multicolumn{1}{c|}{2.49\degree}  &
\multicolumn{1}{|c|}{\bf{3.16}} & \multicolumn{1}{c|}{\bf{1.03\degree}} & \multicolumn{1}{c|}{\bf{2.98}} & \multicolumn{1}{c|}{\bf{1.63\degree}}  &\multicolumn{1}{l}{\small Ours+BA} \\ \cline{1-8} 
\multicolumn{1}{|c|}{64.38} & \multicolumn{1}{c|}{$17.67\degree$} & \multicolumn{1}{c|}{29.38} & \multicolumn{1}{c|}{6.74\degree}  &
\multicolumn{1}{|c|}{82.80} & \multicolumn{1}{c|}{24.10\degree} & \multicolumn{1}{c|}{31.84} & \multicolumn{1}{c|}{15.12\degree}  &\multicolumn{1}{l}{\small COLMAP} \\ \cline{1-8} 
\end{tabular}
\caption{{\bf Loop closure error} (lower is better): for sequences which end at approximately the same location, we can compute the \textit{loop closure error} by duplicating the first image at the end of the sequence, and measuring the error between the two reconstructed views in both position and orientation. This effectively measures the amount of drift over the entire sequence. Position errors are divided by the median baseline to show the deviation in number of frames. The shown configurations are defined in Fig.~\protect{\ref{fig:qualitative}}.}
\vspace{-0.3cm}
\label{tbl:closure_error}
\end{table}

In order to quantify the amount of drift for the complete-building sequences, we copied the first frame in our sequence as the last frame, but ran both Theia and COLMAP without loop closure, and measured the error between the two reconstructed frames.
The quantitative errors in both orientation and position are shown in Table~\ref{tbl:closure_error}.
While adding vanishing point constraints dramatically reduces orientation errors, the positional error is even further reduced by adding the plane constraints.  Performing a final bundle adjustment on this solution even further reduces the error. 

Note that our added constraints do not modify the correspondences, and thus the constraints in bundle adjustment are unchanged. However, our constraints improve the initialization to bundle adjustment, providing reconstructions often much closer to the true solution. This causes bundle adjustment to converge more quickly, and reduces the likelihood of converging to local minima. Since planar constraints are not enforced in bundle adjustment, bundle adjustment could theoretically reintroduce drift, but we have not observed this in any of our experiments. 


\section{Discussion and Conclusions}

\label{sec:conclusions}
In this paper, we have presented a method for efficiently constraining global reconstructions in scenes with man-made structures. We show that the detection and linking of \emph{extended structural features} such as planes and vanishing points, which can span frames that may not have overlapping views of the scene, can be used as powerful additional constraints in structure from motion, enabling the reconstruction of sequences that are particularly susceptible to drift. We demonstrate that these constraints significantly reduce accumulated drift over longer sequences, allowing fast-but-brittle global reconstruction algorithms to rival --- and often surpass --- the accuracy of costly bundle-adjusted incremental systems in scenes with strong structural elements. We present comparisons to state-of-the-art structure-from-motion systems for both synthetic and real-world data.

{\small
\bibliographystyle{bib/ieee}
\bibliography{bib/references}

\begin{thebibliography}{10}\itemsep=-1pt

\bibitem{Agarwal2009}
S.~Agarwal, N.~Snavely, I.~Simon, S.~M. Seitz, and R.~Szeliski.
\newblock Building {Rome} in a day.
\newblock In {\em International Conference on Computer Vision (ICCV)}, pages
  72--79, 2009.

\bibitem{Baillard1999}
C.~Baillard, C.~Schmid, A.~Zisserman, and A.~Fitzgibbon.
\newblock Automatic line matching and {3D} reconstruction of buildings from
  multiple views.
\newblock In {\em ISPRS Conference on Automatic Extraction of GIS Objects from
  Digital Imagery}, volume~32, pages 69--80, 1999.

\bibitem{Bartoli2003}
A.~Bartoli and P.~Sturm.
\newblock Constrained structure and motion from multiple uncalibrated views of
  a piecewise planar scene.
\newblock {\em International Journal of Computer Vision}, 52(1):45--64, 2003.

\bibitem{Bartoli2000}
A.~Bartoli, P.~Sturm, and R.~Horaud.
\newblock {\em A projective framework for structure and motion recovery from
  two views of a piecewise planar scene}.
\newblock PhD thesis, INRIA, 2000.

\bibitem{Bhowmick2014}
B.~Bhowmick, S.~Patra, A.~Chatterjee, V.~M. Govindu, and S.~Banerjee.
\newblock Divide and conquer: Efficient large-scale structure from motion using
  graph partitioning.
\newblock In {\em Asian Conference on Computer Vision (ACCV)}, pages 273--287,
  2014.

\bibitem{Camposeco2015}
F.~Camposeco and M.~Pollefeys.
\newblock Using vanishing points to improve visual-inertial odometry.
\newblock In {\em 2015 IEEE international conference on robotics and automation
  (ICRA)}, pages 5219--5225. IEEE, 2015.

\bibitem{Chatterjee2013}
A.~Chatterjee and V.~Madhav~Govindu.
\newblock Efficient and robust large-scale rotation averaging.
\newblock In {\em Proceedings of the IEEE International Conference on Computer
  Vision}, pages 521--528, 2013.

\bibitem{Cohen2015}
A.~Cohen, T.~Sattler, and M.~Pollefeys.
\newblock Merging the unmatchable: Stitching visually disconnected sfm models.
\newblock In {\em Proceedings of the IEEE International Conference on Computer
  Vision}, pages 2129--2137, 2015.

\bibitem{Cui2015}
Z.~Cui and P.~Tan.
\newblock Global structure-from-motion by similarity averaging.
\newblock In {\em International Conference on Computer Vision (ICCV)}, pages
  864--872, 2015.

\bibitem{Debevec1996}
P.~E. Debevec, C.~J. Taylor, and J.~Malik.
\newblock Modeling and rendering architecture from photographs: A hybrid
  geometry and image-based approach.
\newblock In {\em ACM Transactions on Graphics (Proceedings of SIGGRAPH)},
  volume~96, pages 11--20, 1996.

\bibitem{Engel2018}
J.~Engel, V.~Koltun, and D.~Cremers.
\newblock Direct sparse odometry.
\newblock {\em IEEE Transactions on Pattern Analysis and Machine Intelligence},
  40(3):611--625, 2018.

\bibitem{Engel2014}
J.~Engel, T.~Sch{\"o}ps, and D.~Cremers.
\newblock {LSD-SLAM}: Large-scale direct monocular {SLAM}.
\newblock In {\em European Conference on Computer Vision (ECCV)}, pages
  834--849, 2014.

\bibitem{Faugeras1992}
O.~D. Faugeras, Q.-T. Luong, and S.~J. Maybank.
\newblock Camera self-calibration: Theory and experiments.
\newblock In {\em European Conference on Computer Vision (ECCV)}, pages
  321--334, 1992.

\bibitem{Forster2016}
C.~Forster, Z.~Zhang, M.~Gassner, M.~Werlberger, and D.~Scaramuzza.
\newblock Svo: Semidirect visual odometry for monocular and multicamera
  systems.
\newblock {\em IEEE Transactions on Robotics}, 33(2):249--265, 2016.

\bibitem{Frahm2010}
J.-M. Frahm, P.~Fite-Georgel, D.~Gallup, T.~Johnson, R.~Raguram, C.~Wu, Y.-H.
  Jen, E.~Dunn, B.~Clipp, S.~Lazebnik, et~al.
\newblock Building {Rome} on a cloudless day.
\newblock In {\em European Conference on Computer Vision (ECCV)}, pages
  368--381, 2010.

\bibitem{Fuentes2015}
J.~Fuentes-Pacheco, J.~Ruiz-Ascencio, and J.~M. Rend{\'o}n-Mancha.
\newblock Visual simultaneous localization and mapping: a survey.
\newblock {\em Artificial Intelligence Review}, 43(1):55--81, 2015.

\bibitem{Govindu2001}
V.~M. Govindu.
\newblock Combining two-view constraints for motion estimation.
\newblock In {\em IEEE Conference on Computer Vision and Pattern Recognition
  (CVPR)}, 2001.

\bibitem{Govindu2006}
V.~M. Govindu.
\newblock Robustness in motion averaging.
\newblock In {\em Asian Conference on Computer Vision (ACCV)}, pages 457--466,
  2006.

\bibitem{vonGioi2012}
R.~Grompone~von Gioi, J.~Jakubowicz, J.-M. Morel, and G.~Randall.
\newblock {LSD: a Line Segment Detector}.
\newblock {\em {Image Processing On Line}}, 2:35--55, 2012.

\bibitem{Haralick1989}
R.~M. Haralick, H.~Joo, C.-N. Lee, X.~Zhuang, V.~G. Vaidya, and M.~B. Kim.
\newblock Pose estimation from corresponding point data.
\newblock In {\em Machine Vision for Inspection and Measurement}, pages 1--84.
  Elsevier, 1989.

\bibitem{Harris1988}
C.~Harris and M.~Stephens.
\newblock A combined corner and edge detector.
\newblock In {\em Alvey vision conference}, 1988.

\bibitem{Hartley2003}
R.~Hartley and A.~Zisserman.
\newblock {\em Multiple view geometry in computer vision}.
\newblock Cambridge University Press, 2003.

\bibitem{Hartley1997}
R.~I. Hartley.
\newblock Lines and points in three views and the trifocal tensor.
\newblock {\em International Journal of Computer Vision}, 22(2):125--140, 1997.

\bibitem{Jiang2013}
N.~Jiang, Z.~Cui, and P.~Tan.
\newblock A global linear method for camera pose registration.
\newblock In {\em International Conference on Computer Vision (ICCV)}, pages
  481--488, 2013.

\bibitem{Kosecka2002}
J.~Kosecka and W.~Zhang.
\newblock Video compass.
\newblock In {\em European Conference on Computer Vision (ECCV)}, pages
  476--490, 2002.

\bibitem{Lang1996}
F.~Lang and W.~F{\"o}rstner.
\newblock Surface reconstruction of man-made objects using polymorphic
  mid-level features and generic scene knowledge.
\newblock {\em International Archives of Photogrammetry and Remote Sensing},
  31:415--420, 1996.

\bibitem{Li2018}
H.~Li, J.~Yao, J.-C. Bazin, X.~Lu, Y.~Xing, and K.~Liu.
\newblock A monocular slam system leveraging structural regularity in manhattan
  world.
\newblock In {\em 2018 IEEE International Conference on Robotics and Automation
  (ICRA)}, pages 2518--2525. IEEE, 2018.

\bibitem{LiMouriks-ijrr2013}
M.~Li and A.~I. Mourikis.
\newblock High-precision, consistent {EKF}-based visual-inertial odometry.
\newblock {\em International Journal of Robotics Research}, 32(6):690--711, May
  2013.

\bibitem{Liu2016}
H.~Liu, G.~Zhang, and H.~Bao.
\newblock Robust keyframe-based monocular slam for augmented reality.
\newblock In {\em 2016 IEEE International Symposium on Mixed and Augmented
  Reality (ISMAR)}, pages 1--10. IEEE, 2016.

\bibitem{Lowe1999}
D.~G. Lowe.
\newblock Object recognition from local scale-invariant features.
\newblock In {\em International Conference on Computer Vision (ICCV)},
  volume~2, pages 1150--1157, 1999.

\bibitem{Malis2007}
E.~Malis and M.~Vargas.
\newblock Deeper understanding of the homography decomposition for vision-based
  control.
\newblock Technical Report 6303, INRIA, 2007.

\bibitem{Micusik2017}
B.~Micusik and H.~Wildenauer.
\newblock Structure from motion with line segments under relaxed endpoint
  constraints.
\newblock {\em International Journal of Computer Vision}, 124(1):65--79, 2017.

\bibitem{Moulon2013}
P.~Moulon, P.~Monasse, and R.~Marlet.
\newblock Global fusion of relative motions for robust, accurate and scalable
  structure from motion.
\newblock In {\em International Conference on Computer Vision (ICCV)}, pages
  3248--3255, 2013.

\bibitem{Mur2015}
R.~Mur-Artal, J.~M.~M. Montiel, and J.~D. Tardos.
\newblock {ORB-SLAM}: a versatile and accurate monocular {SLAM} system.
\newblock {\em IEEE Transactions on Robotics}, 31(5):1147--1163, 2015.

\bibitem{Ni2007}
K.~Ni, D.~Steedly, and F.~Dellaert.
\newblock Out-of-core bundle adjustment for large-scale {3D} reconstruction.
\newblock In {\em International Conference on Computer Vision (ICCV)}, 2007.

\bibitem{Nister2004}
D.~Nist{\'e}r.
\newblock An efficient solution to the five-point relative pose problem.
\newblock {\em IEEE Transactions on Pattern Analysis and Machine Intelligence},
  26(6):756--770, 2004.

\bibitem{Nurutdinova2015}
I.~Nurutdinova and A.~Fitzgibbon.
\newblock Towards pointless structure from motion: {3D} reconstruction and
  camera parameters from general {3D} curves.
\newblock In {\em International Conference on Computer Vision (ICCV)}, pages
  2363--2371, 2015.

\bibitem{Ozyesil2015}
O.~Ozyesil and A.~Singer.
\newblock Robust camera location estimation by convex programming.
\newblock In {\em Proceedings of the IEEE Conference on Computer Vision and
  Pattern Recognition}, pages 2674--2683, 2015.

\bibitem{Pollefeys2002}
M.~Pollefeys, F.~Verbiest, and L.~Van~Gool.
\newblock Surviving dominant planes in uncalibrated structure and motion
  recovery.
\newblock In {\em European Conference on Computer Vision (ECCV)}, pages
  837--851, 2002.

\bibitem{Qin2018}
T.~Qin, P.~Li, and S.~Shen.
\newblock Vins-mono: A robust and versatile monocular visual-inertial state
  estimator.
\newblock {\em IEEE Transactions on Robotics}, 34(4):1004--1020, 2018.

\bibitem{Rother2003}
C.~Rother.
\newblock Linear multi-view reconstruction of points, lines, planes and cameras
  using a reference plane.
\newblock In {\em International Conference on Computer Vision (ICCV)}, 2003.

\bibitem{schindler2004atlanta}
G.~Schindler and F.~Dellaert.
\newblock Atlanta world: An expectation maximization framework for simultaneous
  low-level edge grouping and camera calibration in complex man-made
  environments.
\newblock In {\em Proceedings of the 2004 IEEE Computer Society Conference on
  Computer Vision and Pattern Recognition, 2004. CVPR 2004.} IEEE, 2004.

\bibitem{Schonberger2016}
J.~L. Sch\"{o}nberger and J.-M. Frahm.
\newblock Structure-from-motion revisited.
\newblock In {\em IEEE Conference on Computer Vision and Pattern Recognition
  (CVPR)}, 2016.

\bibitem{Schops2014}
T.~{Sch\"{o}ps}, J.~{Engel}, and D.~{Cremers}.
\newblock Semi-dense visual odometry for {AR} on a smartphone.
\newblock In {\em IEEE International Symposium on Mixed and Augmented Reality
  (ISMAR)}, pages 145--150, 2014.

\bibitem{Shariati2019}
A.~Shariati, B.~Pfrommer, and C.~J. Taylor.
\newblock Simultaneous localization and layout model selection in manhattan
  worlds.
\newblock {\em IEEE Robotics and Automation Letters}, 4(2):950--957, 2019.

\bibitem{Sinha2010}
S.~N. Sinha, D.~Steedly, and R.~Szeliski.
\newblock A multi-stage linear approach to structure from motion.
\newblock In {\em ECCV 2010 Workshop on Reconstruction and Modeling of
  Large-Scale 3D Virtual Environments}, 2010.

\bibitem{Snavely2010}
N.~Snavely.
\newblock Bundler: Structure from motion ({SfM}) for unordered image
  collections.
\newblock {\em Code available at http://phototour.cs.washington.edu/bundler/},
  2010.

\bibitem{Snavely2006}
N.~Snavely, S.~M. Seitz, and R.~Szeliski.
\newblock Photo tourism: exploring photo collections in {3D}.
\newblock {\em ACM Transactions on Graphics (Proceedings of SIGGRAPH)},
  25(3):835--846, 2006.

\bibitem{Steedly2003}
D.~Steedly, I.~Essa, and F.~Dellaert.
\newblock Spectral partitioning for structure from motion.
\newblock In {\em International Conference on Computer Vision (ICCV)}, 2003.

\bibitem{stewenius2006recent}
H.~Stewenius, C.~Engels, and D.~Nist{\'e}r.
\newblock Recent developments on direct relative orientation.
\newblock {\em ISPRS Journal of Photogrammetry and Remote Sensing},
  60(4):284--294, 2006.

\bibitem{Straub2014}
J.~Straub, G.~Rosman, O.~Freifeld, J.~J. Leonard, and J.~W. Fisher, III.
\newblock A mixture of {M}anhattan frames: Beyond the {M}anhattan world.
\newblock In {\em IEEE Conference on Computer Vision and Pattern Recognition
  (CVPR)}, 2014.

\bibitem{Sweeney2015}
C.~Sweeney, T.~Hollerer, and M.~Turk.
\newblock Theia: A fast and scalable structure-from-motion library.
\newblock In {\em ACM International Conference on Multimedia}, pages 693--696,
  2015.

\bibitem{Szeliski1998}
R.~Szeliski and P.~H. Torr.
\newblock Geometrically constrained structure from motion: Points on planes.
\newblock In {\em European Workshop on 3D Structure from Multiple Images of
  Large-Scale Environments}, pages 171--186. Springer, 1998.

\bibitem{Torr1997}
P.~H. Torr and A.~Zisserman.
\newblock Robust parameterization and computation of the trifocal tensor.
\newblock {\em Image and Vision Computing}, 15(8):591--605, 1997.

\bibitem{Triggs1999}
B.~Triggs, P.~F. McLauchlan, R.~I. Hartley, and A.~W. Fitzgibbon.
\newblock Bundle adjustment—a modern synthesis.
\newblock In {\em International Workshop on Vision Algorithms}, pages 298--372,
  1999.

\bibitem{Tuytelaars2008}
T.~Tuytelaars, K.~Mikolajczyk, et~al.
\newblock Local invariant feature detectors: a survey.
\newblock {\em Foundations and Trends in Computer Graphics and Vision},
  3(3):177--280, 2008.

\bibitem{Wilson2014}
K.~Wilson and N.~Snavely.
\newblock Robust global translations with {1DSfM}.
\newblock In {\em European Conference on Computer Vision (ECCV)}, pages 61--75,
  2014.

\bibitem{Wu2011}
C.~Wu.
\newblock {VisualSFM}: A visual structure from motion system.
\newblock {\em Code available at http://ccwu.me/vsfm/}, 2011.

\bibitem{Wu2011_CVPR}
C.~Wu, S.~Agarwal, B.~Curless, and S.~M. Seitz.
\newblock Multicore bundle adjustment.
\newblock In {\em IEEE Conference on Computer Vision and Pattern Recognition
  (CVPR)}, pages 3057--3064, 2011.

\bibitem{Yang2019}
S.~Yang and S.~Scherer.
\newblock Monocular object and plane slam in structured environments.
\newblock {\em IEEE Robotics and Automation Letters}, 4(4):3145--3152, 2019.

\bibitem{Zhou2015}
H.~Zhou, D.~Zou, L.~Pei, R.~Ying, P.~Liu, and W.~Yu.
\newblock Structslam: Visual slam with building structure lines.
\newblock {\em IEEE Transactions on Vehicular Technology}, 64(4):1364--1375,
  2015.

\end{thebibliography}
}

\newpage
\appendix
\section*{Appendix (Supplemental Materials)}

 Section~\ref{sec:modifications} provides a description of the modifications that were made to the baseline Theia system in order to support our real-world sequences, which contain degenerate cases such as near-linear camera motions and single planes. Section~\ref{sec:vp-details} contains a more detailed description of our implementation of vanishing point estimation and an analysis of the effects of applying vanishing point constraints on rotational drift. Section~\ref{sec:plane-details} contains details about our implementation of plane fitting in local 3D point clouds. Finally, section~\ref{sec:loop_closure} discusses the effects of enabling explicit loop closure in our real-world sequences and presents a comparison of all our qualitative results with and without loop closure. 
\section{Modifications to Theia}
\label{sec:modifications}
\subsection{Co-linear camera motion}
\label{sec:scale_constraints}
As mentioned in Section~\ref{sec:degenerate}, pairwise constraints are insufficient for certain camera configurations, such as co-linear motion, which is often found in video-based reconstruction. A common tactic for reducing ambiguity and error in the final reconstruction is to verify pairwise estimates among local groups of three frames (triplets) \cite{Jiang2013, Moulon2013}.  
These methods triangulate local 3D points between every pair of the cameras within a triplet and then compare the point depths to establish scale constraints between pairwise translation estimates. 

In order to make Theia robust to co-linear motion, we use a variant of \cite{Moulon2013}, establishing scale constraints between pairs of local pairwise reconstructions and integrating them into Theia's robust position solver \cite{Ozyesil2015}. 

In detail, we first search for triplets as any three frames that have valid three-way pairwise pose estimates. Local 3D points are triangulated for each pairwise estimate in the triplet, using the relative rotations and translations they define. 
Then, relative scales are computed between pairs of these pairwise estimates as the robust ratio between the triangulated point inverse depths.
These scale constraints are added as soft constraints to the previous linear system (Equation~\ref{eqn:global-unweighted}) as:
\begin{equation}
  E_{\mbox{scale}}(\{s_{ij}, s_{kl}\}) = W_s(ij,kl) * \left(r_{ij\rightarrow{}kl}s_{ij} - s_{kl} \right),
    \label{eqn:scale_constraints}
\end{equation} where $s_{ij}, s_{kl}$ are the global unknown scale values for pairwise local reconstructions $i\rightarrow{}j$ and $k\rightarrow{}l$, and $r_{ij\rightarrow{}kl}$ is the previously computed robust ratio. The strength of the added constraint is weighted by $W_s(ij,kl)$, defined as 
\begin{equation}
  W_s(ij,kl) = \max\left(\frac{N_{\mbox{pts}}(ij, jk)}{N_{max}},1\right)
    \label{eqn:scale_weighting}
\end{equation} where $N_{\mbox{pts}}(ij,jk)$ is the number of shared tracks with valid triangulations\footnote{as defined by Theia's standard two-view triangulation code, which includes triangulation angle, among other metrics                                                 } between pairwise reconstructions $i\rightarrow{}j$ and $j\rightarrow{}k$, and $N_{max}$ is the number of points at which the weight saturates (we use $N_{max} = 500$).

\subsection{Rotation-only camera motion and planar scenes}
\label{sec:homography}
At the core of Theia's global SfM pipeline lies its pairwise relative pose estimates, which are used as constraints in both in the global rotation and position solvers. These pairwise relative poses are estimated using the five-point algorithm \cite{stewenius2006recent, Nister2004}, which is known to produce unreliable estimates in cases of purely rotational motion or entirely planar scenes. Since our sequences largely consist of close-up captures of building facades (planes), we incorporate a secondary pipeline for pairwise relative pose estimation. 

In addition to the usual five-point estimation, we estimate a homography between the matching feature points. If it is determined that a majority of points are considered inliers to the homography, the estimated homography is decomposed using \cite{Malis2007}, resulting in four candidate rotation-translation transformation pairs. These putative transformations are subsequently filtered by first discarding those which triangulate points with mostly negative depths, and then the transformation with the smallest reprojection error is retained. The remainder of the pose estimation pipeline remains unchanged.

\section{Vanishing point estimation and integration}
\label{sec:vp-details}

In order to obtain a drift-free set of rotation estimates, we first compute for each frame (wherever possible) a vertical vanishing point and one or more
horizontal vanishing points.
We first detect line segments using the LSD line segment detector \cite{vonGioi2012}. 
To fit vanishing points, we use the general expectation maximization 
approach of \cite{Kosecka2002}. In the expectation stage, line segments are softly associated to vanishing points; in the maximization stage, vanishing points are fit to line segments by solving a weighted least squares problem on the unit sphere. Between iterations, we merge vanishing points that become sufficiently close, and purge vanishing points with low evidence.

This approach requires a method to find an initial set of vanishing points. For this, we start by finding vertical vanishing point candidates from lines that are close to vertical in the image. For each candidate vertical direction, we compute the corresponding horizon line. Candidate horizontal vanishing points are found by intersecting image lines with the horizon line. We divide the horizon line into equal-angle bins and select candidate vanishing points from peaks in the resulting histogram. We can also obtain a candidate set of vanishing points from the previous frame, if any. To select among the candidate sets, we choose the set which maximizes the total length of lines associated with a vanishing point.

Once the vanishing points have been found in frame $i$, we want to estimate a global rotation $\Rmat_i^{\mbox{vp}}$, which maps one of the Atlanta world horizontal directions to the dominant horizontal direction in the frame. As mentioned earlier, our method is tailored for
continuous video sequences where motion between frames is small, so the vanishing point associations can be chained through consecutive frames. For most cases, the vanishing directions are consistent across consecutive frames, since the distribution of edge orientations in the image remains approximately constant, and thus the orientation can be computed as the relative rotation from the dominant horizontal axis. However, in certain cases, for example when turning a corner, the dominant vanishing direction may change. In these cases, we use the pairwise pose estimate from the last frame with valid vanishing directions to verify the association. This verification can result in either keeping the estimated vanishing point as-is, applying a 90 degree rotation (eg. turning the corner of a building), or begin tracking a new set of horizontal vanishing directions altogether (eg. when turning a non-orthogonal building corner, like in an Atlanta-world). We keep whichever association produces the most consistent vanishing point association, and invalidate any associations with errors greater than 10 degrees. While this same verification process could be used to associate vanishing points in arbitrary frames, we still limit our method to sequential data, since we found empirically that lifting this assumption, i.e. performing this association between all frames with pairwise estimates, often results in spurious VP associations.

Once the local coordinate frames have been estimated, they need to be integrated into the rotation averaging step.
As mentioned in Section~\ref{subsec:vanishing points}, 
these constraints are weighted using a regularization parameter $\lvp$ and a per-frame vanishing point weighting function $\Wvp$,
which we define as 
\begin{equation}
  \Wvp(i) = \max (1, 1 - \frac{\Delta \theta}{\theta_{max}})
\end{equation}
where $\Delta \theta$ is the incremental rotation between frames $i-1$ and $i$ and $\theta_{max}$ is the largest tolerated incremental rotation. We use $\theta_{max}=5$ degrees for all experiments. This weighting function makes the rotation estimates more robust to significant outliers in the vanishing point estimates, by lowering the weight of consecutive frames when they are very different. 
\begin{figure}
  \hfill\includegraphics[width=\linewidth]{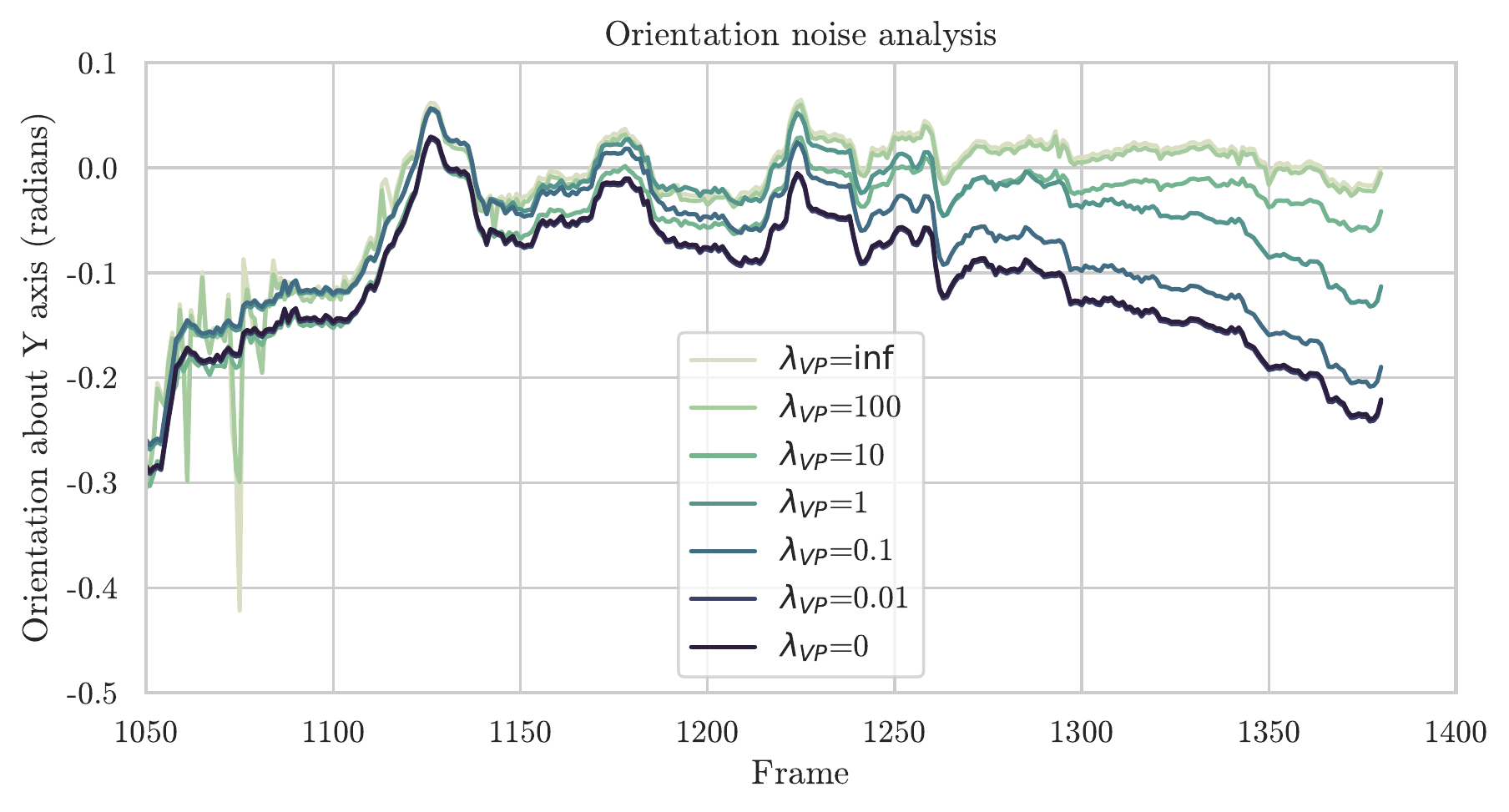}
  \caption{\small{A comparison of the reconstructed camera orientations for the \texttt{MORE\char`_HALF} sequence  using different values of $\lambda_{vp}$.
  We see the reconstructed orientations suffer from significant low-frequency drift when no vanishing point constraints are applied ($\lambda_{vp} = 0$). This is equivalent to result shown in Figure~8a of the main paper. Increasing the weight of the vanishing point constraints causes the low frequency drift to decrease ($\lambda = 0.1, 1, 10$), but too large values ($\lambda = 100$) will introduce high-frequency noise from the vanishing point estimates into the reconstructed orientations, seen as spikes in the curve.}}
  \label{fig:noise_comparison}
\end{figure}

In order to determine a suitable value for $\lvp$, we could estimate the variance of the local inter-frame rotation estimates by comparing them to the rotation-averaged solution, and the variance in the vanishing point estimates by comparing incremental rotations between them to the rotation averaged differences.  Instead, we took the simpler approach of just setting a value empirically by observing the rotation averaged plots for various values of $\lvp$.
Fig.~\ref{fig:noise_comparison} shows the orientation estimates for a representative set of frames (the last 322 frames of the \texttt{MORE\char`_HALF} sequence) reconstructed using various values of $\lvp$.
Notice that with weak vanishing point constraints ($\lvp = 0$ or $0.01$), the results drift significantly from the global orientations given by the vanishing points. 
On the other hand, the vanishing point estimates are occasionally wrong, such as the large spike seen around frame 1075. 
We found experimentally that setting $\lvp = 10$ gave us good results for all of the sequences that we tested.\begin{figure}
  \hfill\includegraphics[width=\linewidth]{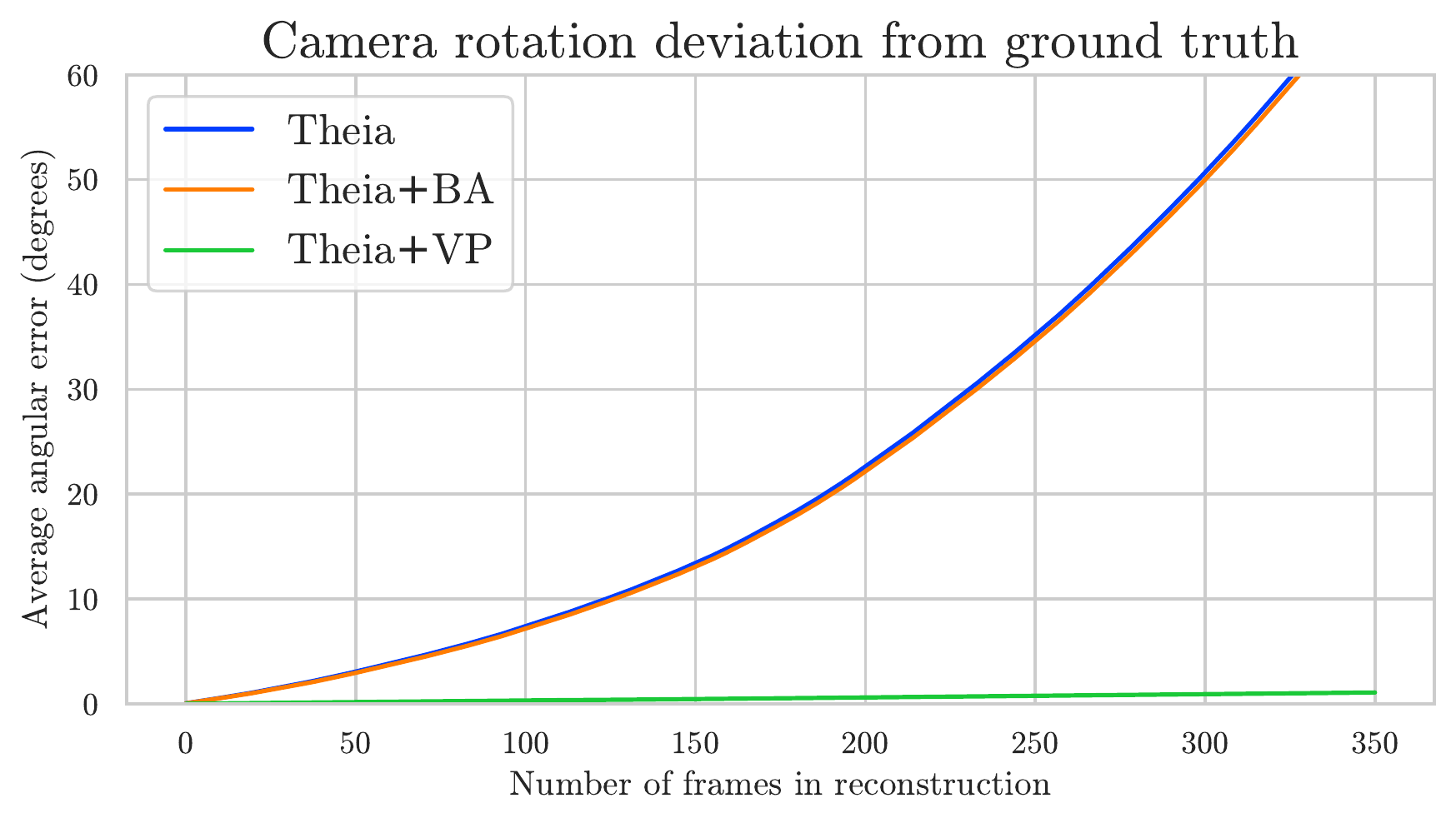}
\caption{{\bf Quantitative evaluation (rotational drift)}: 
 A comparison of rotational drift using different combinations of Theia with global bundle adjustment (BA) and added vanishing point constraints (VP).
  We see that the added constraints practically remove the observed rotational drift, and without the added constraints, even bundle adjustment has difficulty converging on the correct solution.
  In order to be robust to pose errors in individual frames, we align the reconstructions to ground truth by estimating a similarity transformation. }
\label{fig:rotation_error}
\end{figure}

Using this value, we show in Fig.~\ref{fig:rotation_error} that adding vanishing point constraints to rotation averaging results in the elimination of rotational drift in our synthetic sequence.

\section{Plane fitting details}
\label{sec:plane-details}

As mentioned in Section \ref{subsec:planes}, our plane fitting process begins by using pairwise pose estimates to triangulate feature point matches into local 3D point clouds. We then use plane sweeps to detect planes along the three orthogonal vanishing directions associated with the local reconstruction's base frame. In this section, we describe in more detail the process through which we perform these plane sweeps.

We begin by projecting the 3D points along the direction we plan to sweep (i.e. performing a dot product of the direction vector and the 3D point). The resulting projected points are then sorted  by depth (distance along the sweep direction from the pairwise origin) producing a one-dimensional histogram. A moving window algorithm is used to detect peaks in this histogram. Peaks correspond to sets of coplanar points along the sweep direction. In peak detection, we use a window size proportional to the predicted scale of the pairwise reconstruction (as defined in \ref{sec:scale_constraints}) and also weight points by their inverse depths, since closer planes provide more useful constraints. We only retain peaks corresponding to at least $N_{min}$ points (in practice, we use $N_{min}$=3). Finally, each peak is used to parameterize a local plane $\pi$, with $\nhat_\pi^{ij}$ equal to the sweep direction, and $d_{ij}^\pi$ equal to the depth of the peak.

\section{Loop closed results}
\label{sec:loop_closure}
Loop closure, in the context of an SfM system like Theia or COLMAP, only has significance when the matching strategy relies on temporal proximity, i.e. frames are only matched to nearby frames in the video sequence. In these cases, feature matching may not be performed between the first and last frames in a sequence, even if they observe the same parts of the scene. Loop closure addresses this by adding pairwise pose estimates between frames which may not have otherwise been matched, thus adding constraints to the rotation and position estimation stages. These pairs of frames can be identified by a number of strategies, including vocabulary trees and spatial proximity in incremental reconstruction.

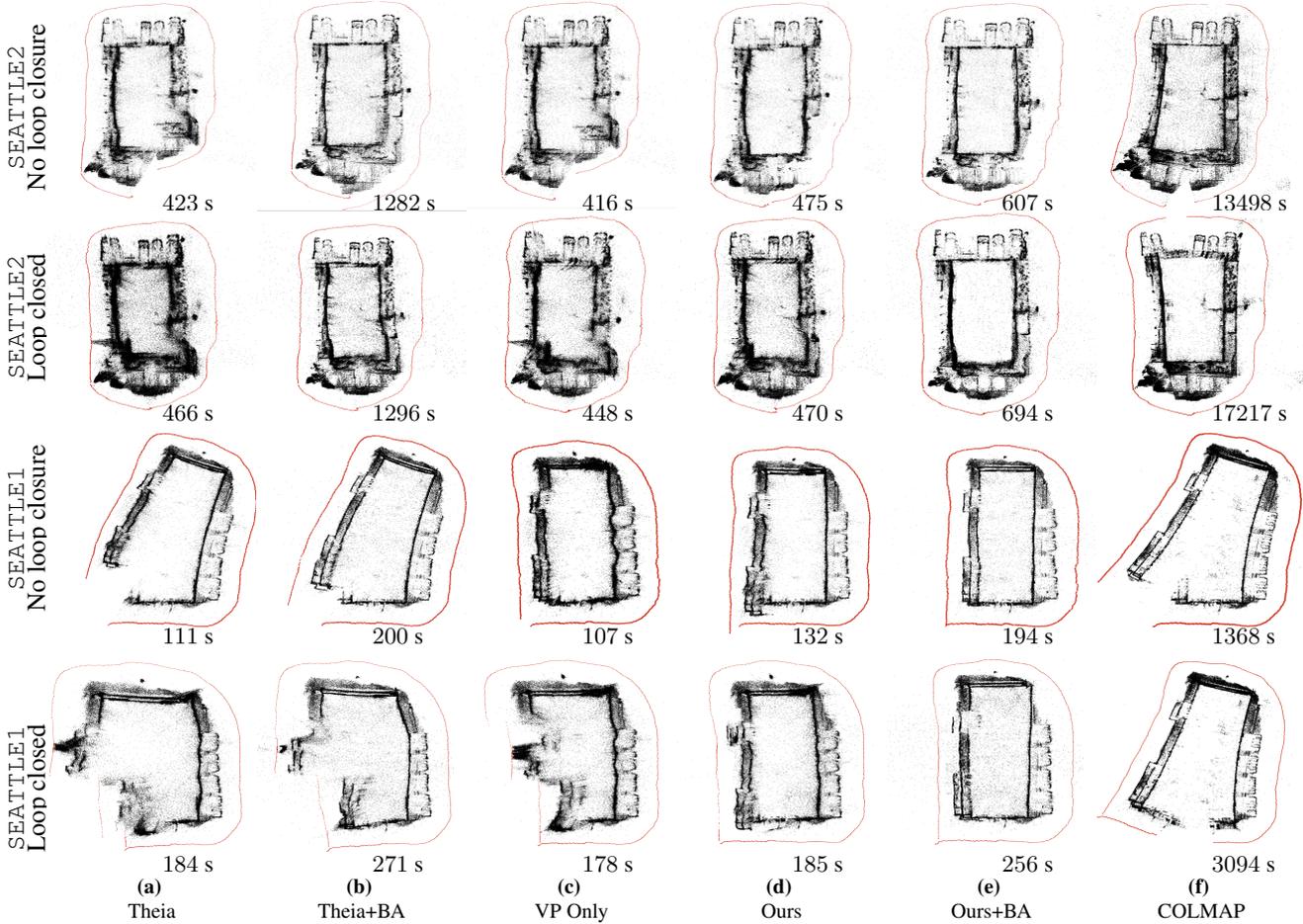
\begin{figure*}
    \centering 
\begin{subfigure}{0.015\textwidth}
  \rotatebox{90}{\texttt{SEATTLE2}}\
\end{subfigure}\hfil 
\begin{subfigure}{0.015\textwidth}
  \rotatebox{90}{No loop closure}\
\end{subfigure}\hfil 
\begin{subfigure}{0.1616\textwidth}
  \begin{overpic}[height=\linewidth,width=\linewidth]{figures/images/golden_eagle2/vanilla.png}
    \put (55,0) {\small $423$ s}
  \end{overpic}
\end{subfigure}\hfil 
\begin{subfigure}{0.1616\textwidth}
  \begin{overpic}[height=\linewidth,width=\linewidth]{figures/images/golden_eagle2/vanilla_bundle.png}
    \put (55,0) {\small $1282$ s}
  \end{overpic}
\end{subfigure}\hfil 
\begin{subfigure}{0.1616\textwidth}
  \begin{overpic}[height=\linewidth,width=\linewidth]{figures/images/golden_eagle2/vp.png}
    \put (55,0) {\small $416$ s}
  \end{overpic}
\end{subfigure}\hfil 
\begin{subfigure}{0.1616\textwidth}
  \begin{overpic}[height=\linewidth,width=\linewidth]{figures/images/golden_eagle2/plane.png}
    \put (55,0) {\small $475$ s}
  \end{overpic}
\end{subfigure}\hfil 
\begin{subfigure}{0.1616\textwidth}
  \begin{overpic}[height=\linewidth,width=\linewidth]{figures/images/golden_eagle2/plane_bundle.png}
    \put (55,0) {\small $607$ s}
  \end{overpic}
\end{subfigure}\hfil 
\begin{subfigure}{0.1616\textwidth}
  \begin{overpic}[height=\linewidth,width=\linewidth]{figures/images/golden_eagle2/colmap.png}
    \put (55,0) {\small $13498$ s}
  \end{overpic}
\end{subfigure}
\begin{subfigure}{0.015\textwidth}
  \rotatebox{90}{\texttt{SEATTLE2}}\
\end{subfigure}\hfil 
\begin{subfigure}{0.015\textwidth}
  \rotatebox{90}{Loop closed}\
\end{subfigure}\hfil 
\begin{subfigure}{0.1616\textwidth}
  \begin{overpic}[height=\linewidth,width=\linewidth]{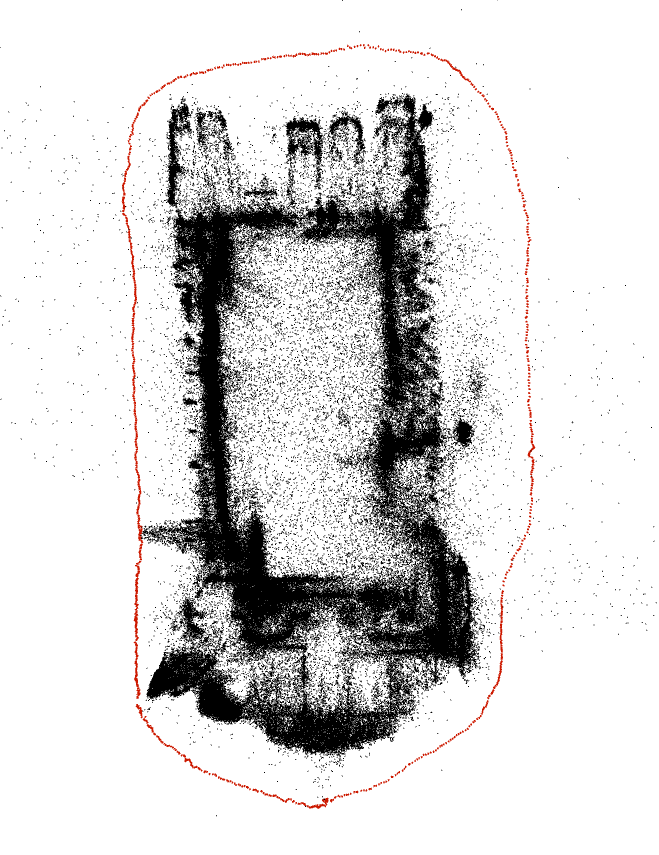}
    \put (55,0) {\small $466$ s}
  \end{overpic}
\end{subfigure}\hfil 
\begin{subfigure}{0.1616\textwidth}
  \begin{overpic}[height=\linewidth,width=\linewidth]{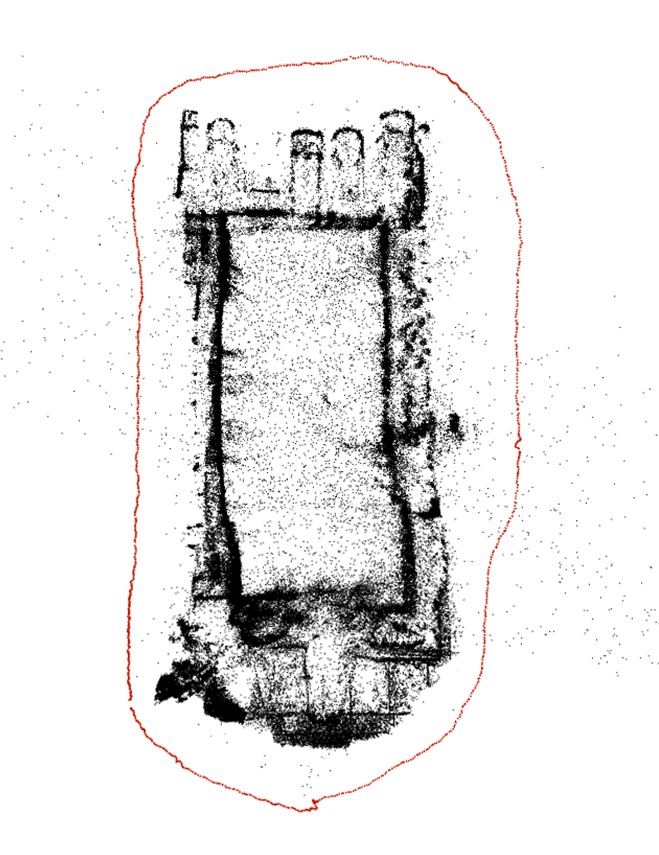}
    \put (55,0) {\small $1296$ s}
  \end{overpic}
\end{subfigure}\hfil 
\begin{subfigure}{0.1616\textwidth}
  \begin{overpic}[height=\linewidth,width=\linewidth]{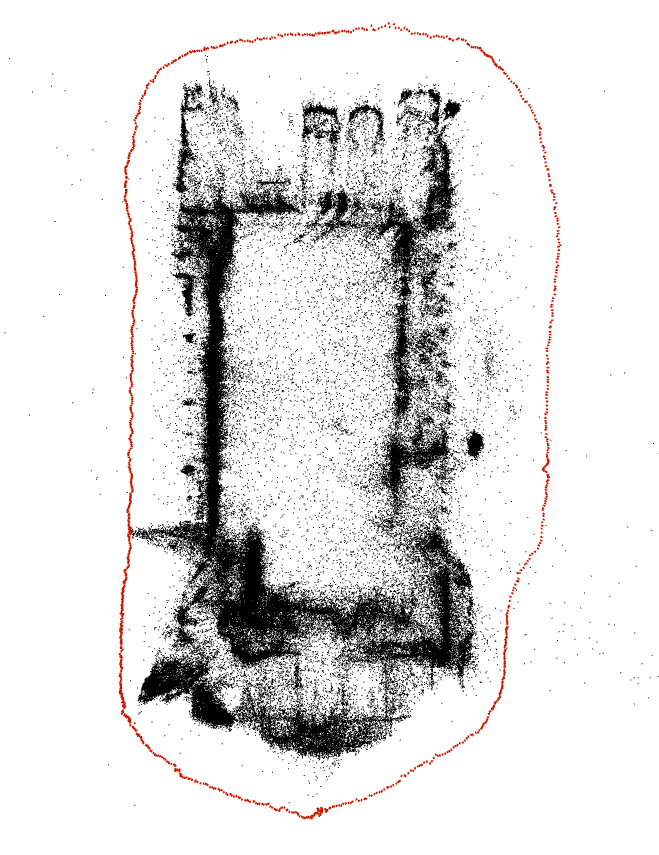}
    \put (55,0) {\small $448$ s}
  \end{overpic}
\end{subfigure}\hfil 
\begin{subfigure}{0.1616\textwidth}
  \begin{overpic}[height=\linewidth,width=\linewidth]{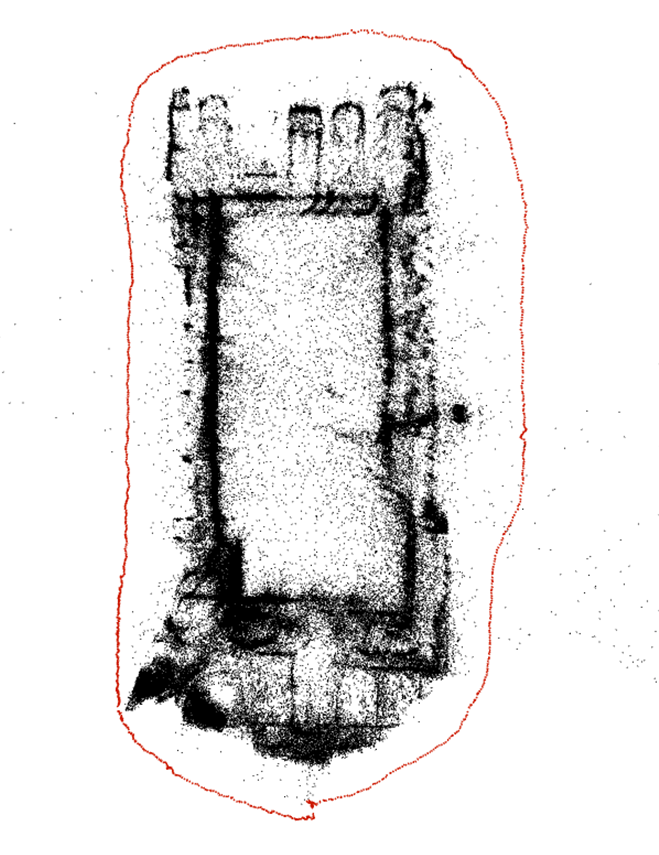}
    \put (55,0) {\small $470$ s}
  \end{overpic}
\end{subfigure}\hfil 
\begin{subfigure}{0.1616\textwidth}
  \begin{overpic}[height=\linewidth,width=\linewidth]{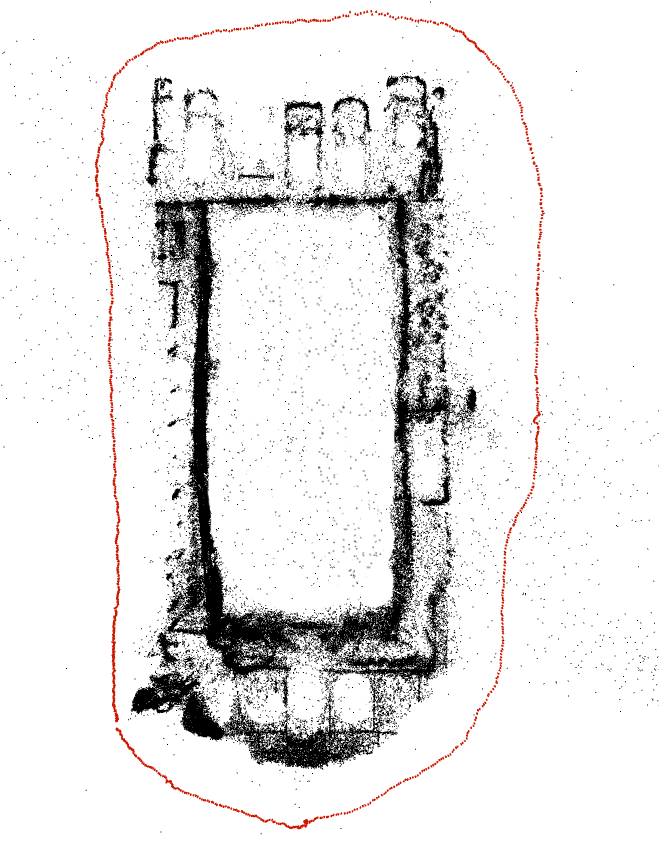}
    \put (55,0) {\small $694$ s}
  \end{overpic}
\end{subfigure}\hfil 
\begin{subfigure}{0.1616\textwidth}
  \begin{overpic}[height=\linewidth,width=\linewidth]{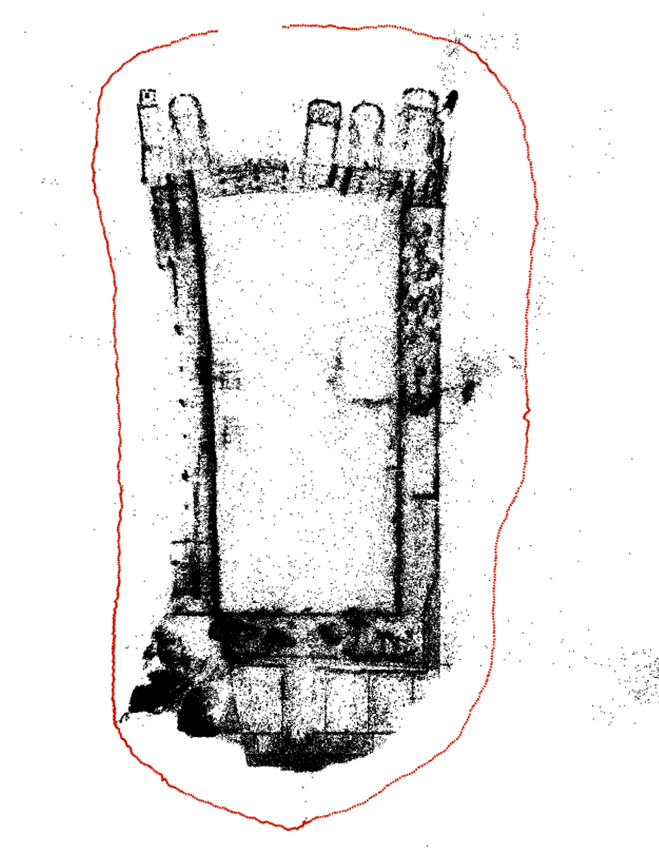}
    \put (55,0) {\small $17217$ s}
  \end{overpic}
\end{subfigure}
\medskip
\begin{subfigure}{0.015\textwidth}
  \rotatebox{90}{\texttt{SEATTLE1}}\
\end{subfigure}\hfil 
\begin{subfigure}{0.015\textwidth}
  \rotatebox{90}{No loop closure}\
\end{subfigure}\hfil 
\begin{subfigure}{0.1616\textwidth}
  \begin{overpic}[height=\linewidth,width=\linewidth]{figures/images/cortina/vanilla.png}
    \put (55,-5) {\small $111$ s}
  \end{overpic}
\end{subfigure}\hfil 
\begin{subfigure}{0.1616\textwidth}
  \begin{overpic}[height=\linewidth,width=\linewidth]{figures/images/cortina/vanilla_bundle.png}
    \put (55,-5) {\small $200$ s}
  \end{overpic}
\end{subfigure}\hfil 
\begin{subfigure}{0.1616\textwidth}
  \begin{overpic}[height=\linewidth,width=\linewidth]{figures/images/cortina/vp.png}
    \put (55,-5) {\small $107$ s}
  \end{overpic}
\end{subfigure}\hfil 
\begin{subfigure}{0.1616\textwidth}
  \begin{overpic}[height=\linewidth,width=\linewidth]{figures/images/cortina/plane.png}
    \put (55,-5) {\small $132$ s}
  \end{overpic}
\end{subfigure}\hfil 
\begin{subfigure}{0.1616\textwidth}
  \begin{overpic}[height=\linewidth,width=\linewidth]{figures/images/cortina/plane_bundle.png}
    \put (55,-5) {\small $194$ s}
  \end{overpic}
\end{subfigure}\hfil 
\begin{subfigure}{0.1616\textwidth}
  \begin{overpic}[height=\linewidth,width=\linewidth]{figures/images/cortina/colmap.png}
    \put (55,-5) {\small $1368$ s}
  \end{overpic}
\end{subfigure}
\medskip
\begin{subfigure}{0.015\textwidth}
  \rotatebox{90}{\texttt{SEATTLE1}}\
\end{subfigure}\hfil 
\begin{subfigure}{0.015\textwidth}
  \rotatebox{90}{Loop closed}\
\end{subfigure}\hfil 
\begin{subfigure}{0.1616\textwidth}
  \begin{overpic}[height=\linewidth,width=0.938\linewidth]{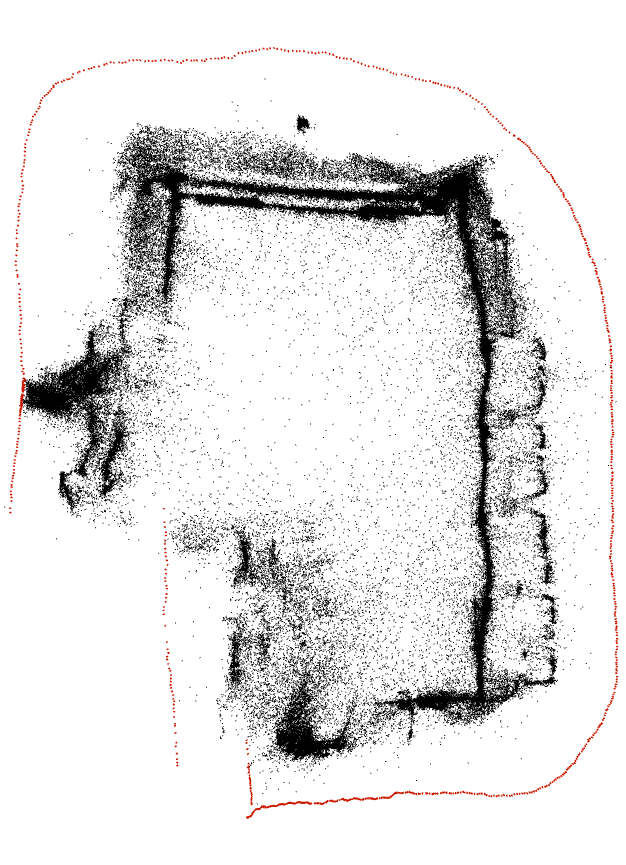}
    \put (55,-5) {\small $184$ s}
  \end{overpic}
    \captionsetup{labelfont=bf}
  \caption{\\Theia}
\end{subfigure}\hfil 
\begin{subfigure}{0.1616\textwidth}
  \begin{overpic}[height=\linewidth,width=0.938\linewidth]{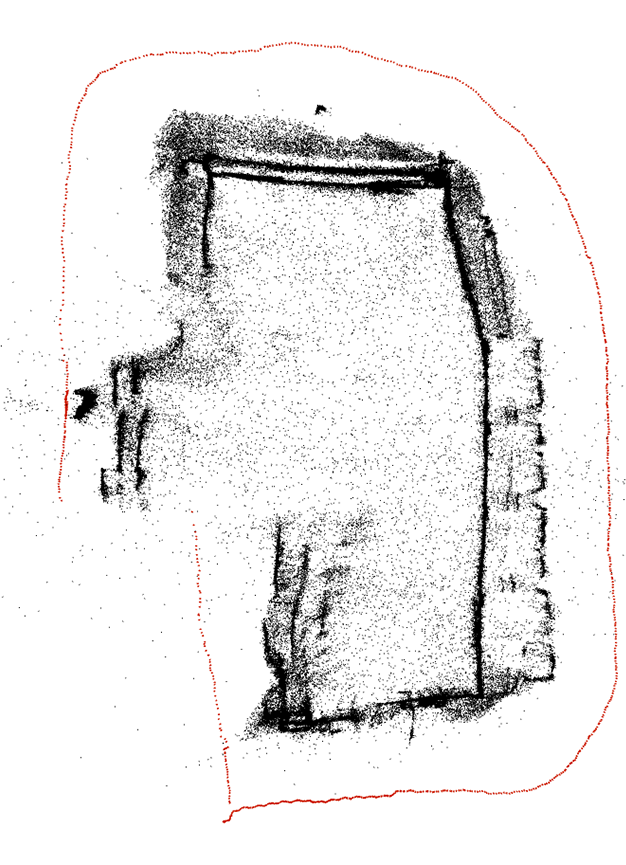}
    \put (55,-5) {\small $271$ s}
  \end{overpic}
    \captionsetup{labelfont=bf}
  \caption{\\Theia+BA}
\end{subfigure}\hfil 
\begin{subfigure}{0.1616\textwidth}
  \begin{overpic}[height=\linewidth,width=0.938\linewidth]{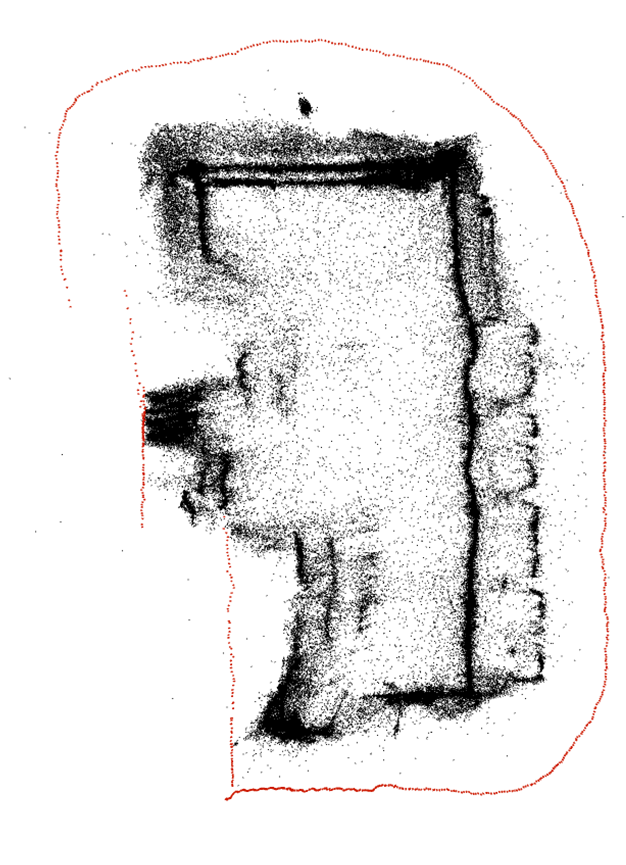}
    \put (55,-5) {\small $178$ s}
  \end{overpic}
    \captionsetup{labelfont=bf}
  \caption{\\VP Only}
\end{subfigure}\hfil 
\begin{subfigure}{0.1616\textwidth}
  \begin{overpic}[height=\linewidth,width=0.938\linewidth]{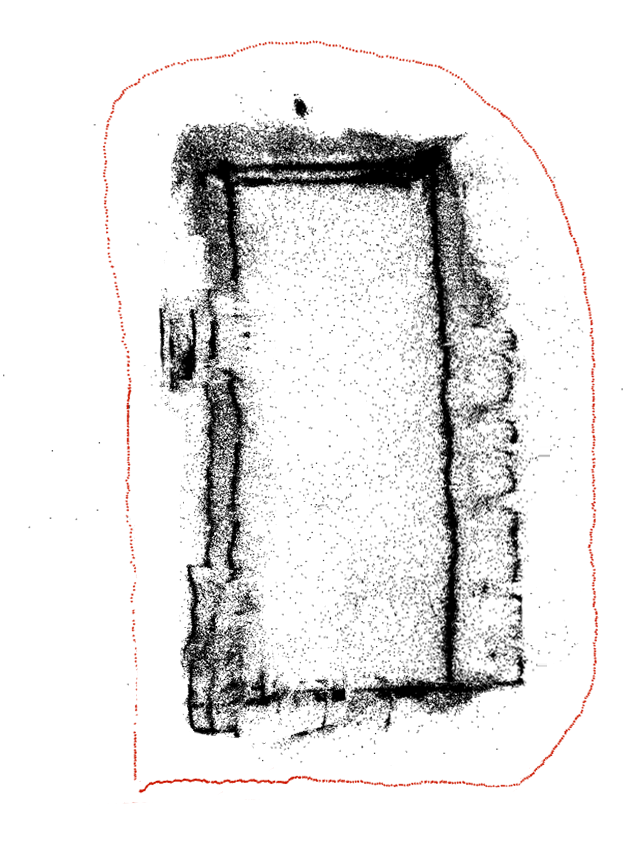}
    \put (55,-5) {\small $185$ s}
  \end{overpic}
    \captionsetup{labelfont=bf}
  \caption{\\Ours}
\end{subfigure}\hfil 
\begin{subfigure}{0.1616\textwidth}
  \begin{overpic}[height=\linewidth,width=0.938\linewidth]{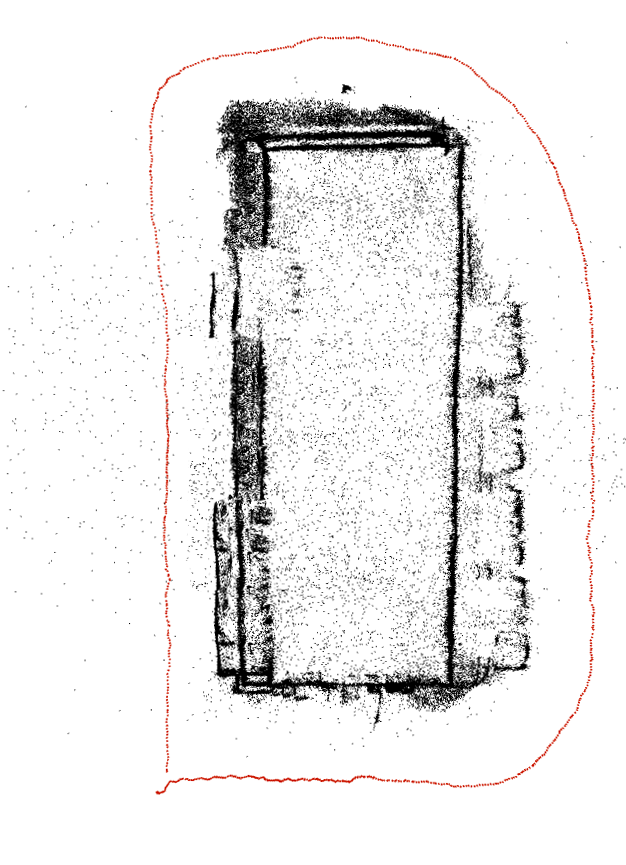}
    \put (55,-5) {\small $256$ s}
  \end{overpic}
    \captionsetup{labelfont=bf}
  \caption{\\Ours+BA}
\end{subfigure}\hfil 
\begin{subfigure}{0.1616\textwidth}
  \begin{overpic}[height=\linewidth,width=0.938\linewidth]{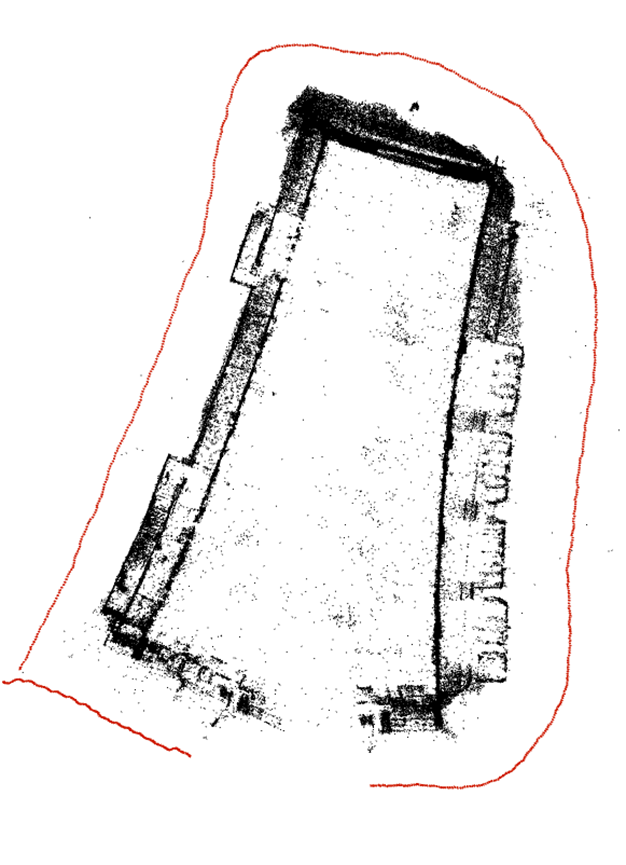}
    \put (55,-5) {\small $3094$ s}
  \end{overpic}
    \captionsetup{labelfont=bf}
  \caption{\\COLMAP}
\end{subfigure}

\caption{{\bf Loop closure comparison}: top-down views of the real-world datasets using different reconstruction methods:
{\bf (a)} Theia without added structural constraints or global bundle adjustment (BA),
{\bf (b)} Theia without added constraints but with BA,
{\bf (c)} Theia with added vanishing point constraints (VP), but no planar constraints (PC) or BA
{\bf (d)} Theia with both VP and PC, but without BA
{\bf (e)} Theia with VP, PC, and BA,
{\bf (f)} COLMAP, with BA.
Each sequence is shown twice, both with and without automatic loop closure. We can see that in the \texttt{SEATTLE2} sequence, while both the addition of loop closure and bundle adjustment produce more reasonable looking reconstructions, the point clouds still contains significant errors, as seen in the bent walls of the building. In the top row, we see that Theia struggles to fully reconstruct the structure of the bottom right corner of the building without the help of our structural constraints. In the second row, we see that even though Theia+BA produces a connected loop, the bottom half of the building facade is significantly bent. As can be seen in column (e), our added constraints resolve both these errors. In the \texttt{SEATTLE1} sequence, the addition of loop closure adds significant errors in the form of multiple discontinuities in camera position and noisy reconstructed structure. Even so, our method is able to recover a reconstruction similar in quality to the non-loop-closed variant. In both sequences, COLMAP's loop closed reconstruction simply moves the trajectory discontinuities seen in the non-loop-closed version to a different part of the sequence, whereas our results produce straight walls and continuous camera trajectories. The remaining two closed-loop sequences (\texttt{SEATTLE3}, \texttt{ATLANTA1}) contain significant repetitive structure resulting in large numbers of spurious loop closure matches. As a result, a reasonable reconstruction was not achieved through any of the shown configurations with automatic loop closure enabled. \vspace{-0.4cm}}
\label{fig:qualitative_loop}
\end{figure*}

When dealing with buildings that can be circumnavigated, one might think that automatic loop closure would resolve all drift error. As can be seen in Fig.~~\ref{fig:qualitative_loop}, this is not the case for our sequences. For two of our sequences (\texttt{SEATTLE3}, \texttt{ATLANTA1}) the addition of automatic loop closure caused the reconstruction to collapse, due to spurious loop-closure matches of repetitive structures. For the remainder of the sequences, while introducing loop closure does indeed pull the camera centers of the first and last frames closer to one another, visible errors emerge elsewhere.

Figure~\ref{fig:qualitative_loop} also demonstrates that our added structural constraints perform similarly even when loop closure is enabled. In fact, when comparing to the addition of loop closure, we see that our added constraints are both more performant and accurate.

For our experiments in Sec.~\ref{sec:real_world}, we do not employ loop closure, in order to more visibly demonstrate the effects of drift (for the qualitative experiments) and to be able to quantify drift (for the quantitative experiment in Table~\ref{tbl:closure_error}). After all, loop closure is not always an option, as not all sequences return to the same viewpoint, and even when they do, loop closure techniques can sometimes fail to find a match, especially when closure overlap is minimal. Furthermore, as we see in our experiments, while loop closure constraints encourage the trajectory endpoints to align, they do not guarantee the elimination of all errors resulting from drift.


\end{document}